\documentclass[runningheads]{llncs}
\usepackage{graphicx}
\usepackage[table]{xcolor}
\usepackage{adjustbox}

\usepackage{wrapfig}
\usepackage{tikz}
\usepackage{comment}
\usepackage{amsmath,amssymb} 
\usepackage{color}
\usepackage{algorithm,algpseudocode}
\usepackage{booktabs} 
\usepackage{multirow}
\newlength{\Oldarrayrulewidth}

\newcommand{\etal}{\textit{et al}. }
\newcommand{\ie}{\textit{i}.\textit{e}., }
 
\usepackage[accsupp]{axessibility}  

\usepackage[bookmarksnumbered=true]{hyperref} 

\hypersetup{
     colorlinks = true,
     linkcolor = black,
     anchorcolor = black,
     citecolor = black,
     filecolor = black,
     urlcolor = purple
     }

\begin{document}
\pagestyle{headings}
\mainmatter
\def\ECCVSubNumber{7345}  

\title{Exploring Lottery Ticket Hypothesis in \\ Spiking Neural Networks} 


\author{Youngeun Kim, Yuhang Li, Hyoungseob Park, Yeshwanth Venkatesha,\\ Ruokai Yin, and Priyadarshini Panda}
\institute{
Department of Electrical Engineering\\
Yale University\\
New Haven, CT, USA\\
{\tt \{youngeun.kim, yuhang.li, hyoungseob.park, yeshwanth.venkatesha, ruokai.yin, priya.panda\}@yale.edu}
}

\titlerunning{Lottery Ticket Hypothesis for SNNs}
\authorrunning{Y. Kim, Y. Li, H. Park, V. Yeshwanth, R. Yin, P. Panda.}
\maketitle

\begin{abstract}

Spiking Neural Networks (SNNs) have recently emerged as a new generation of low-power deep neural networks, which is suitable to be implemented on low-power mobile/edge devices.
As such devices have limited memory storage, neural pruning on SNNs has been widely explored in recent years.
Most existing SNN pruning works focus on shallow SNNs (2$\sim$6 layers), however, deeper SNNs ($\ge$16 layers) are proposed by state-of-the-art SNN works, which is difficult to be compatible with the current SNN pruning work.
To scale up a pruning technique towards deep SNNs, we investigate Lottery Ticket Hypothesis (LTH) which states that dense networks contain smaller subnetworks (\ie winning tickets) that achieve comparable performance to the dense networks. Our studies on LTH reveal that the winning tickets consistently exist in deep SNNs across various datasets and architectures, providing up to $97\%$ sparsity without huge performance  degradation. However, the iterative searching process of LTH brings a huge training computational cost when combined with the multiple timesteps of SNNs. To alleviate such heavy searching cost, we propose Early-Time (ET) ticket where we find the important weight connectivity from a smaller number of timesteps.
The proposed ET ticket can be seamlessly combined with a common pruning techniques for finding winning tickets,
such as Iterative Magnitude Pruning (IMP) and Early-Bird (EB) tickets.
Our experiment results show that the proposed ET ticket reduces search time by up to $38\%$ compared to  IMP or EB methods. 
Code is available at \href{https://github.com/Intelligent-Computing-Lab-Yale/Exploring-Lottery-Ticket-Hypothesis-in-SNNs}{Github}. 

\keywords{Spiking Neural Networks, Neural Network Pruning, Lottery Ticket Hypothesis, Neuromorphic Computing}
\end{abstract}

\section{Introduction}
Spiking Neural Networks (SNNs) \cite{roy2019towards,christensen20222022,wu2018spatio,wu2019direct,kundu2021hire,fang2021deep,wu2022brain,kim2021visual} have gained significant attention as promising low-power alternative to Artificial Neural Networks (ANNs).
Inspired by the biological neuron, SNNs process visual information through discrete spikes over multiple timesteps. 
This event-driven behavior of SNNs brings huge energy-efficiency, therefore they are suitable to be implemented on low-power neuromorphic chips \cite{akopyan2015truenorth,davies2018loihi,furber2014spinnaker,orchard2021efficient} which compute spikes in an asynchronous manner.
However, as such devices have limited memory storage, neural pruning can be one of the essential techniques 
by reducing memory usage for weight parameters, thus promoting the practical deployment.

Accordingly, researchers have made certain progress on the pruning technique for SNNs.
Neftci \etal  \cite{neftci2016stochastic} and Rathi \etal  \cite{rathi2018stdp}
prune weight connections of SNNs using a predefined threshold value.
Guo \etal \cite{guo2020unsupervised} propose an unsupervised online adaptive weight pruning algorithm that dynamically removes non-critical weights over time.
Moreover, Shi \etal \cite{shi2019soft} present a soft-pruning method where both weight connections and a pruning mask are trained during training.
Recently, Deng \etal \cite{deng2021comprehensive}  adapt ADMM optimization tool with sparsity regularization to compress SNNs.
Chen \etal \cite{chen2021pruning} propose a gradient-based rewiring method for pruning, where weight values and connections are jointly optimized.
However, although existing SNN pruning works significantly increase weight sparsity,  they focus on a shallow architecture such as 2-layer MLP \cite{neftci2016stochastic,rathi2018stdp,guo2020unsupervised} or 6$\sim$7 convolutional layers \cite{deng2021comprehensive,chen2021pruning}.
Such pruning techniques are difficult to scale up to the recent state-of-the-art deep SNN architectures where the number of parameters and network depth is scaled up \cite{li2021free,li2022converting,fang2021deep,rathi2021diet,zheng2020going}.

\begin{figure}[t]
\begin{center}
\def\arraystretch{0.5}
\begin{tabular}{@{}c@{\hskip 0.1\linewidth}c@{}c}
\includegraphics[width=0.34\linewidth]{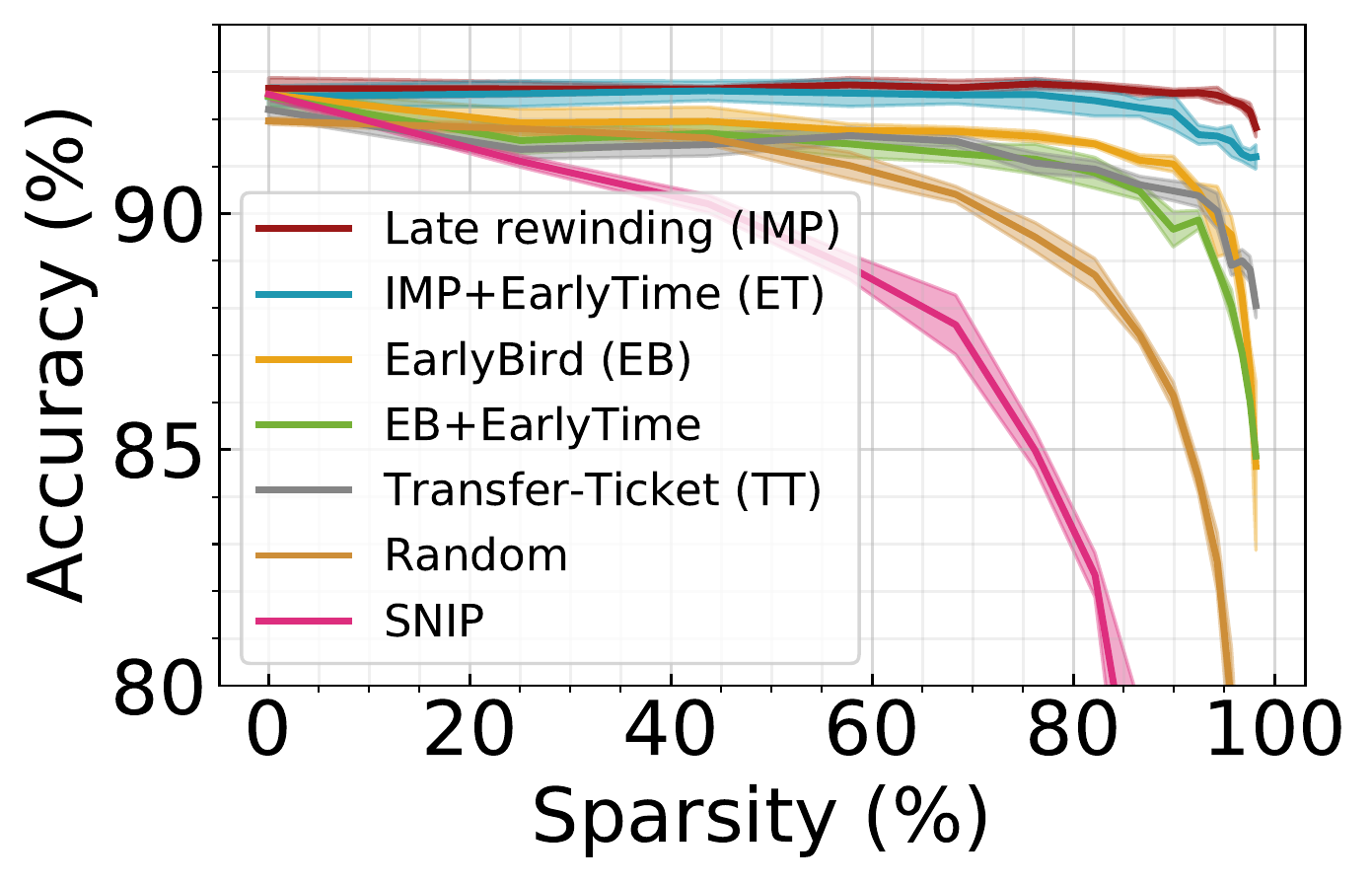} &
\includegraphics[width=0.34\linewidth]{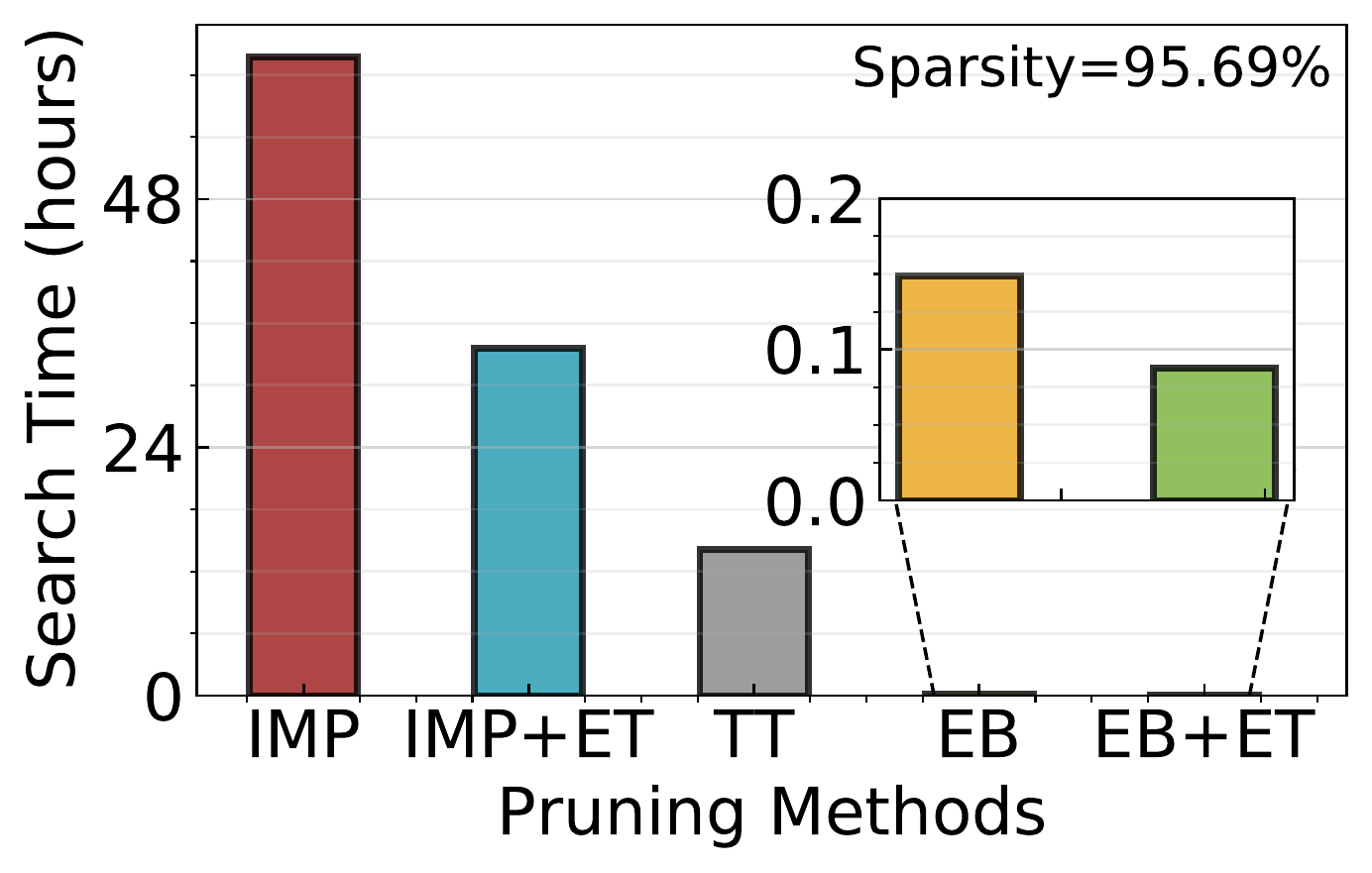}
\\
{ (a) Accuracy} & { (b)  Search Time} \\
\end{tabular}
\caption{Accuracy and Search time comparison of various pruning methods on SNNs including Iterative Magnitude Pruning (IMP) \cite{frankle2018lottery}, Early-Bird (EB) ticket \cite{you2019drawing}, Early-Time (ET) ticket (ours), Transferred winning ticket from ANN (TT), Random pruning, and SNIP \cite{lee2018snip}. We use VGG16 on the CIFAR10 dataset and show mean/standard deviation from 5 random runs.
}
\label{fig:intro:overall_vgg16_cifar10}
\end{center}
\end{figure}

In this paper, we explore sparse deep SNNs based on the recently proposed Lottery Ticket Hypothesis (LTH) \cite{frankle2018lottery}.
They assert that an over-parameterized neural network contains sparse sub-networks that achieve a similar or even better accuracy than the original dense networks.
The discovered sub-networks and their corresponding initialization parameters are referred as \textit{winning tickets}.
Based on LTH, a line of works successfully have shown the existence of winning tickets across various tasks such as standard recognition task \cite{you2019drawing,frankle2019stabilizing,girish2021lottery}, reinforcement learning \cite{vischer2021lottery,yu2019playing}, natural language processing \cite{brix2020successfully,chen2020lottery,movva2020dissecting}, and generative model \cite{kalibhat2020winning}.
 Along the same line, our primary research objective is to investigate the existence of winning tickets in deep SNNs which has a different type of neuronal dynamics from the common ANNs.

Furthermore, applying LTH on SNNs poses a practical challenge.
In general, finding winning tickets requires \textbf{Iterative Magnitude Pruning (IMP)} where a network becomes sparse by repetitive initialization-training-pruning operations \cite{frankle2018lottery}.
Such iterative training process goes much slower with SNNs where multiple feedforward steps (\ie timesteps) are required. 
To make LTH with SNNs  more practical, we explore several techniques for reducing search costs.
We first investigate \textbf{Early-Bird (EB) ticket} phenomenon \cite{you2019drawing} that states the sub-networks can be discovered in the early-training phase.
We find that SNNs contain EB tickets across various architectures  and  datasets.
Moreover, SNNs convey information through multiple timesteps, which provides a new dimension for computational cost reduction.
Focusing on such temporal property, we propose \textbf{Early-Time (ET) ticket} phenomenon: winning tickets can be drawn from the network trained from a smaller number of timesteps.
Thus, during the search process, SNNs use a smaller number of timesteps, which can significantly reduce search costs. 
As ET ticket is a temporal-crafted method, our proposed ET ticket can be combined with IMP \cite{frankle2018lottery} and EB tickets \cite{you2019drawing}.
Furthermore, we also explore whether \textbf{ANN winning ticket can be transferred to SNN} since search cost at ANN is much cheaper than SNN.
Finally, we examine \textbf{pruning at initialization} method on SNNs, \ie  SNIP \cite{lee2018snip}, which finds the winning tickets from backward gradients at initialization.
In Fig. \ref{fig:intro:overall_vgg16_cifar10}, we compare the accuracy and  search time of the above-mentioned pruning methods.

In summary, we explore LTH for SNNs by conducting extensive experiments on two representative deep architectures, \ie VGG16 \cite{simonyan2014very} and ResNet19 \cite{he2016deep}, on  four public datasets including SVHN~\cite{netzer2011reading}, Fashion-MNIST~\cite{xiao2017fashion}, CIFAR10~\cite{krizhevsky2009learning}
and CIFAR100~\cite{krizhevsky2009learning}. Our key observations are as follows: 
\begin{itemize}
\item We confirm that Lottery Ticket Hypothesis is valid for SNNs.

\item We found that IMP \cite{frankle2018lottery} discovers winning tickets with up to 97\% sparsity. However, IMP requires over 50 hours on GPU to find sparse ($\ge95\%$) SNNs.
\item EB tickets \cite{you2019drawing} discover sparse SNNs in 1 hour on GPUs, which  reduces search cost significantly compared to IMP. Unfortunately, they fail to detect winning tickets over 90\% sparsity.
 \item Applying ET tickets to both IMP and EB significantly reduces search time by up to $41\%$ while showing $\le 1\%$ accuracy drop at less than $95\%$ sparsity.
  \item  Winning ticket obtained from ANN can be transferable at less than $90\%$ sparsity. However, huge accuracy drop is incurred  at high sparsity levels ($\ge95\%$), especially for complex datasets such as CIFAR100.
  \item Pruning at initialization method \cite{lee2018snip} fails to discover winning tickets in SNNs owing to the non-differentiability of spiking neurons.
\end{itemize}

\section{Related Work}

\subsection{Spiking Neural Networks}

Spiking Neural Networks (SNNs) transfer binary and asynchronous information through networks in a low-power manner
\cite{roy2019towards,christensen20222022,comsa2020temporal,mostafa2017supervised,schuman2022opportunities,li2021free,venkatesha2021federated,yao2021temporal}.
The major difference of SNNs from standard ANNs is using a Leak-Integrate-and-Fire (LIF) neuron  \cite{izhikevich2003simple} as a non-linear activation.
The LIF neuron accumulates incoming spikes in membrane potential and generates an output spike when the neuron has a higher membrane potential than a firing threshold.
Such integrate-and-fire behavior brings a non-differentiable input-output transfer function where standard backpropagation is difficult to be applied \cite{neftci2019surrogate}.
Recent SNN works circumvent non-differentiable backpropagation problem by defining a surrogate function for LIF neurons when calculating backward gradients \cite{lee2016training,lee2020enabling,neftci2019surrogate,shrestha2018slayer,wu2018spatio,wu2021training,li2021differentiable,kim2022neural,wu2020progressive}. Our work is also based on a gradient backpropagation method with a surrogate function (details are provided in Supplementary A).
The gradient backpropagation methods enable SNNs to have deeper architectures.  
For example, adding Batch Normalization (BN) \cite{ioffe2015batch} to SNNs \cite{ledinauskas2020training,kim2020revisiting,zheng2020going} improves the accuracy with deeper architectures such as VGG16 and ResNet19.
Also, Fang \etal \cite{fang2021deep} revisit deep
residual connection for SNNs, showing higher performance can be achieved by adding more layers.
Although the recent state-of-the-art SNN architecture goes deeper \cite{zheng2020going,fang2021deep,deng2022temporal}, pruning for such networks has not been explored.
We assert that showing the existence of winning tickets in deep SNNs brings a practical advantage to resource-constrained neuromorphic chips and edge devices.

\subsection{Lottery Ticket Hypothesis}

Pruning has been actively explored in recent decades, which compresses a huge model size of the deep neural networks while maintaining its original performance \cite{han2015learning,han2016dsd,wen2016learning,liu2018rethinking,li2016pruning}.
In the same line of thought, Frankle  \& Carbin \cite{frankle2018lottery} present Lottery Ticket Hypothesis (LTH) which states that an over-parameterized neural network contains sparse sub-networks with similar or even better accuracy than the original dense networks.
They search winning tickets by iterative magnitude pruning (IMP). 
Although IMP methods \cite{zhou2019deconstructing,bai2022dual,ding2022audio,burkholz2021existence,liu2021deep} provide higher performance compared to existing pruning methods, such iterative training-pruning-retraining operations require a huge training cost.
To address this, a line of work \cite{mehta2019sparse,morcos2019one,desai2019evaluating,chen2020lottery} discovers the existence of transferable winning tickets from the source dataset and successfully transfers it to the target dataset, thus eliminating search cost.
Further, You \etal \cite{you2019drawing} introduce \textit{early bird ticket} hypothesis where they conjecture winning tickets can be achieved in the early training phase, reducing the cost for training till convergence.
Recently, Zhang \etal \cite{zhang2021efficient} discover winning tickets with a carefully selected subset of training data, called pruning-aware critical set.
To completely eliminate training costs, 
several works \cite{lee2018snip,wang2020picking} present searching algorithms from initialized networks, which finds winning tickets without training.
Unfortunately, such techniques do not show comparable performance with the original IMP methods, thus mainstream LTH leverages IMP as a pruning scheme \cite{girish2021lottery,zhang2021efficient,morcos2019one}.
Based on IMP technique, researchers found the existence of LTH in various applications including visual recognition tasks \cite{girish2021lottery}, natural language processing \cite{brix2020successfully,chen2020lottery,movva2020dissecting}, reinforcement learning \cite{vischer2021lottery,yu2019playing}, generative model \cite{kalibhat2020winning}, low-cost neural network ensembling \cite{liu2021deep}, and improving robustness \cite{chen2022sparsity}.
Although LTH has been actively explored in ANN domain, LTH for SNNs is rarely studied.
It is worth mentioning that  Martinelli \etal \cite{martinelli2020spiking} apply LTH to two-layer SNN on voice activity detection task. 
Different from the previous work, our work shows the existence of winning tickets in much deeper networks such as VGG16 and ResNet19, which shows state-of-the-art performance on the image recognition task.
We also explore Early-Bird ticket \cite{you2019drawing}, SNIP \cite{lee2018snip}, transferability of winning tickets from ANN, and propose a new concept of winning tickets in the temporal dimension.

\begin{figure}[t]
\begin{center}
\def\arraystretch{0.5}
\begin{tabular}{@{}c@{}c@{}c}
\includegraphics[width=0.31\linewidth]{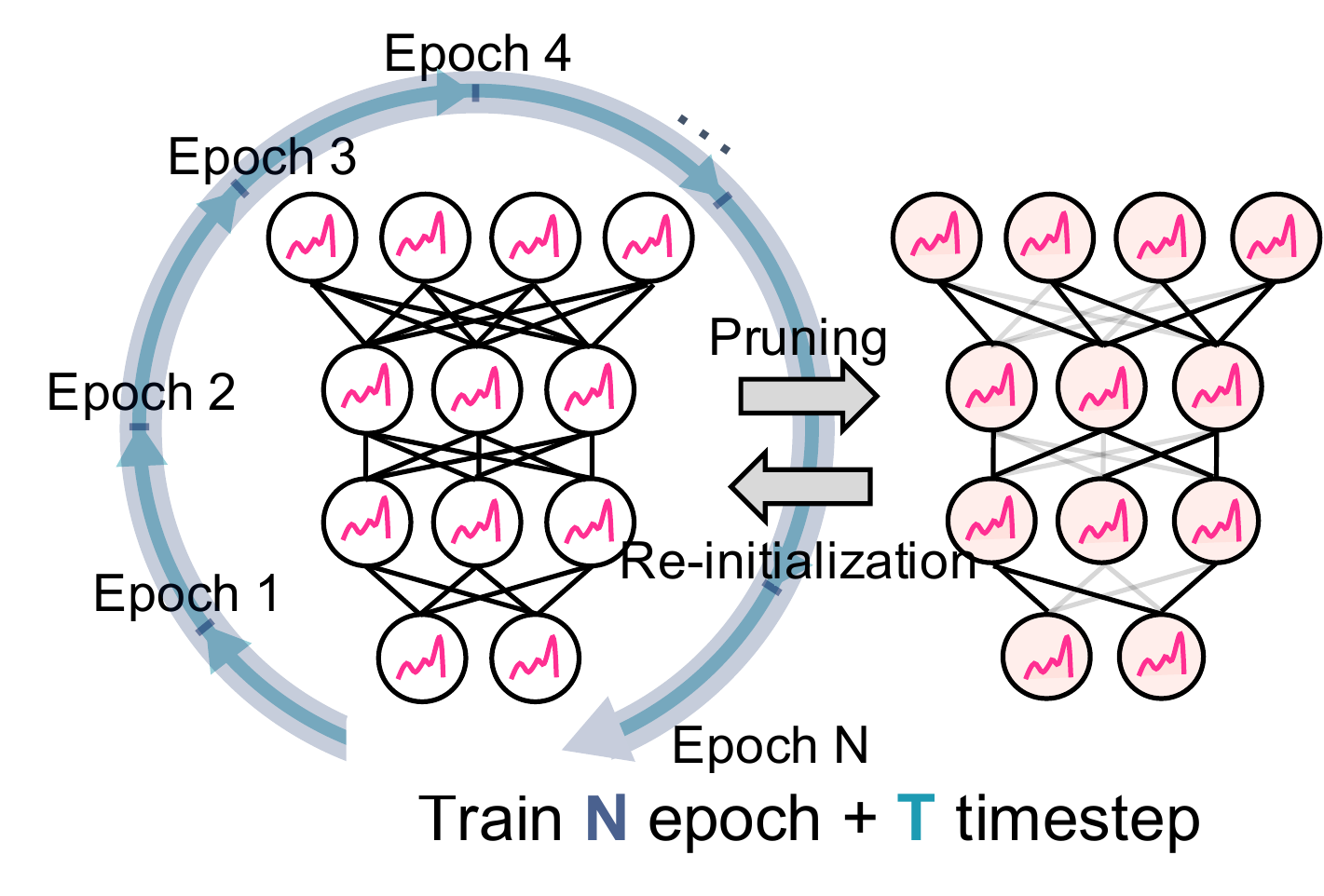} &
\includegraphics[width=0.31\linewidth]{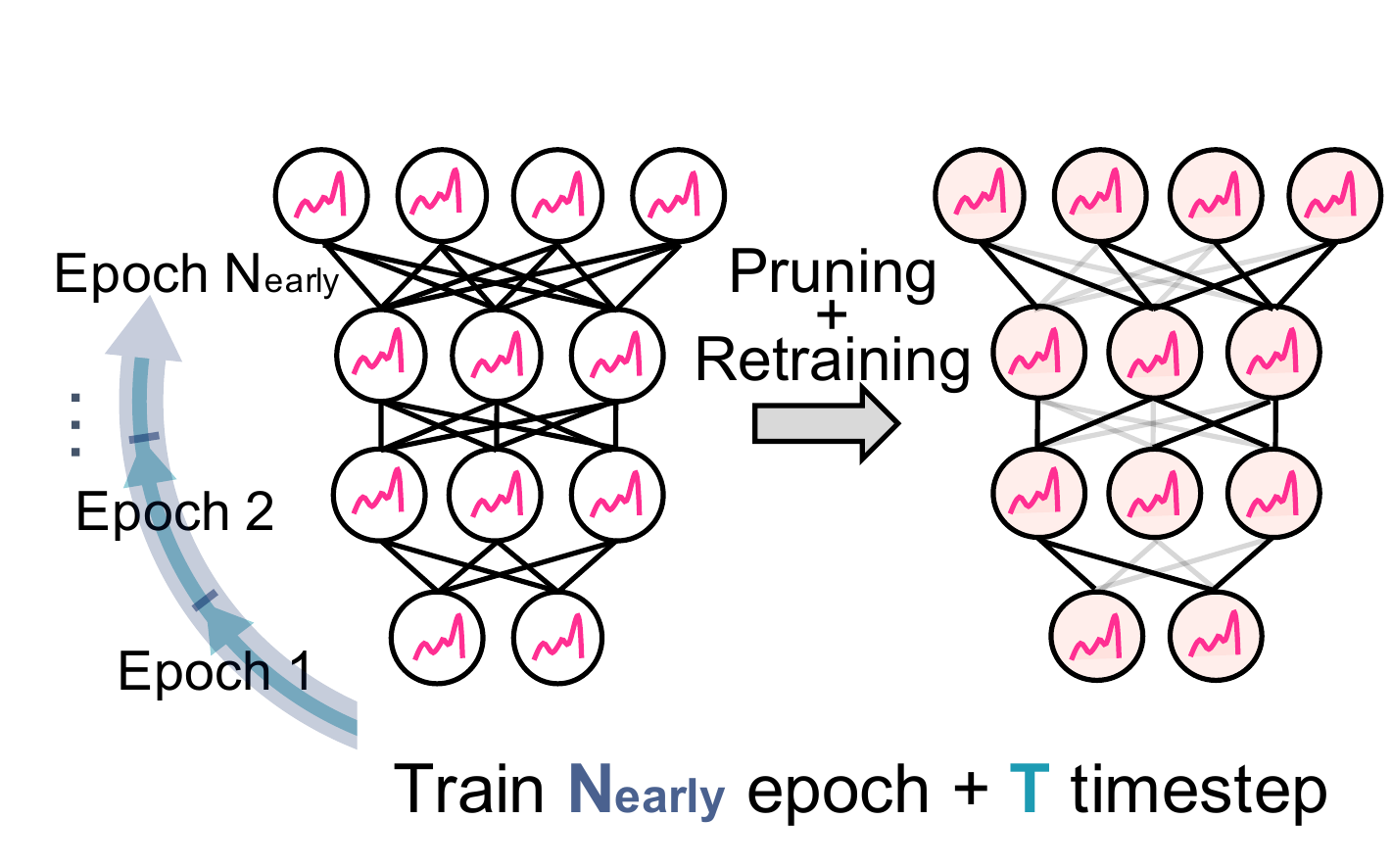} &
\includegraphics[width=0.31\linewidth]{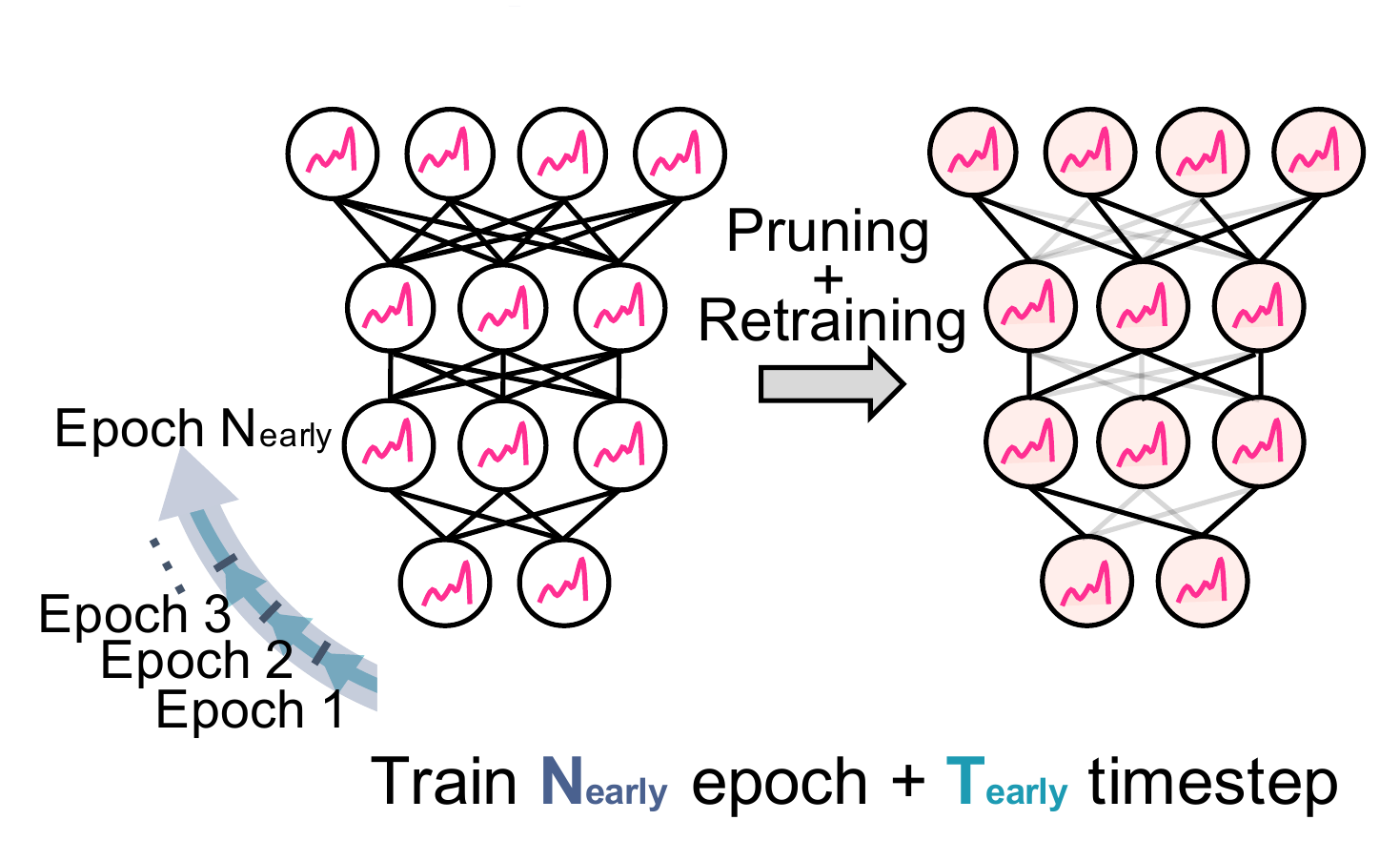}
\\
{ (a) Iterative Magnitude Pruning} & { (b) Early-Bird Ticket} & 
{ (c)  Early-Time Ticket}
\end{tabular}
\caption{Illustration of the concept of Iterative Magnitude Pruning (IMP), Early-Bird (EB) ticket, and the proposed Early-Time (ET) ticket applied to EB ticket. Our ET ticket reduces search cost for winning tickets by using a smaller number of timesteps during the search process. Note, ET can be applied to both IMP and EB, here we only illustrate ET with EB.
}
\label{fig:method:concept_illustration}
\end{center}
\end{figure}

\section{Drawing Winning Tickets from SNN}
In this section, we present the details of pruning methods based on LTH, explored in our experiments.
We first introduce the LTH~\cite{frankle2018lottery} and Early-Bird (EB) Ticket~\cite{you2019drawing}.
Then, we propose Early-Time (ET) Tickets where we reduce search cost in the temporal dimension of SNNs.
We illustrate the overall search process of each method in Fig. \ref{fig:method:concept_illustration}.

\subsection{Lottery Ticket Hypothesis}
In LTH  \cite{frankle2018lottery}, winning tickets are discovered by iterative magnitude pruning (IMP).
The whole pruning process of LTH goes through $K$ iterations for the target pruning ratio $p^K\%$.
Consider a randomly initialized dense network $f(x;\theta)$ where $\theta \in \mathbb{R}^{n}$ is the network parameter weights.
For the first iteration, initialized network $f(x;\theta)$ is trained till convergence, then mask $m_{1}\in \{0, 1\}^{n}$ is generated by removing $p\%$ lowest absolute value weights parameters.
Given a pruning mask $m_1$, we can define subnetworks $f(x;\theta \odot m_1)$ by removing some connections.
For the next iteration, we reinitialize the network with $\theta \odot m_{1}$, and prune p\% weights when the network is trained to convergence.
This pruning process is repeated for $K$ iterations.
In our experiments, we set $p$ and $K$ to $25\%$ and $15$, respectively.
Also, Frankle \etal \cite{frankle2019stabilizing} present \textit{Late Rewinding}, which rewinds the network to the weights at epoch $i$ rather than initialization.
This enables IMP to discover the winning ticket with less performance drop in a high sparsity regime by providing a more stable starting point.
We found that \textit{Late Rewinding} shows better performance than the original IMP in deep SNNs (see Supplementary B).
Throughout our paper, we apply \textit{Late Rewinding} to IMP for experiments where we rewind the network to epoch $20$.

\subsection{Early-Bird Tickets}

Using IMP for finding lottery ticket incurs huge computational costs.
To address this, You \etal \cite{you2019drawing} propose an efficient pruning method, called Early-bird (EB) tickets, where they show winning tickets can be discovered at an early training epoch.
Specifically, at searching iteration $k$, they obtain a mask $m_{k}$ and measure the mask difference between the current mask $m_{k}$ and the previous masks within time window $q$, \ie $m_{k-1}, m_{k-2}, ..., m_{k-q}$. 
If the maximum difference is less than hyperparamter $\tau$, they stop training and use $m_{k}$ as an EB ticket.
As searching winning tickets in SNN takes a longer time than ANN, we explore the existence of EB tickets in SNNs. 
In our experiments, we set $q$ and $\tau$ to $5$ and $0.02$, respectively. Although the original EB tickets use channel pruning, we use unstructured weight pruning for finding EB tickets in order to achieve a similar sparsity level with other pruning methods used in our experiments.

\subsection{Early-Time Tickets}

Even though EB tickets significantly reduce searching time for winning tickets.
For SNNs, one image is passed to a network through multiple timesteps, which provides a new dimension for computational cost reduction.
We ask: 
\textit{can we find the important weight connectivity for the SNN trained with timestep $T$ from the SNN trained with shorter timestep $T'<T$?}

\begin{figure}[t]
\begin{center}
\def\arraystretch{0.5}
\begin{tabular}{@{}c@{}c@{}c@{}c@{}c@{}c}
\includegraphics[width=0.21\linewidth]{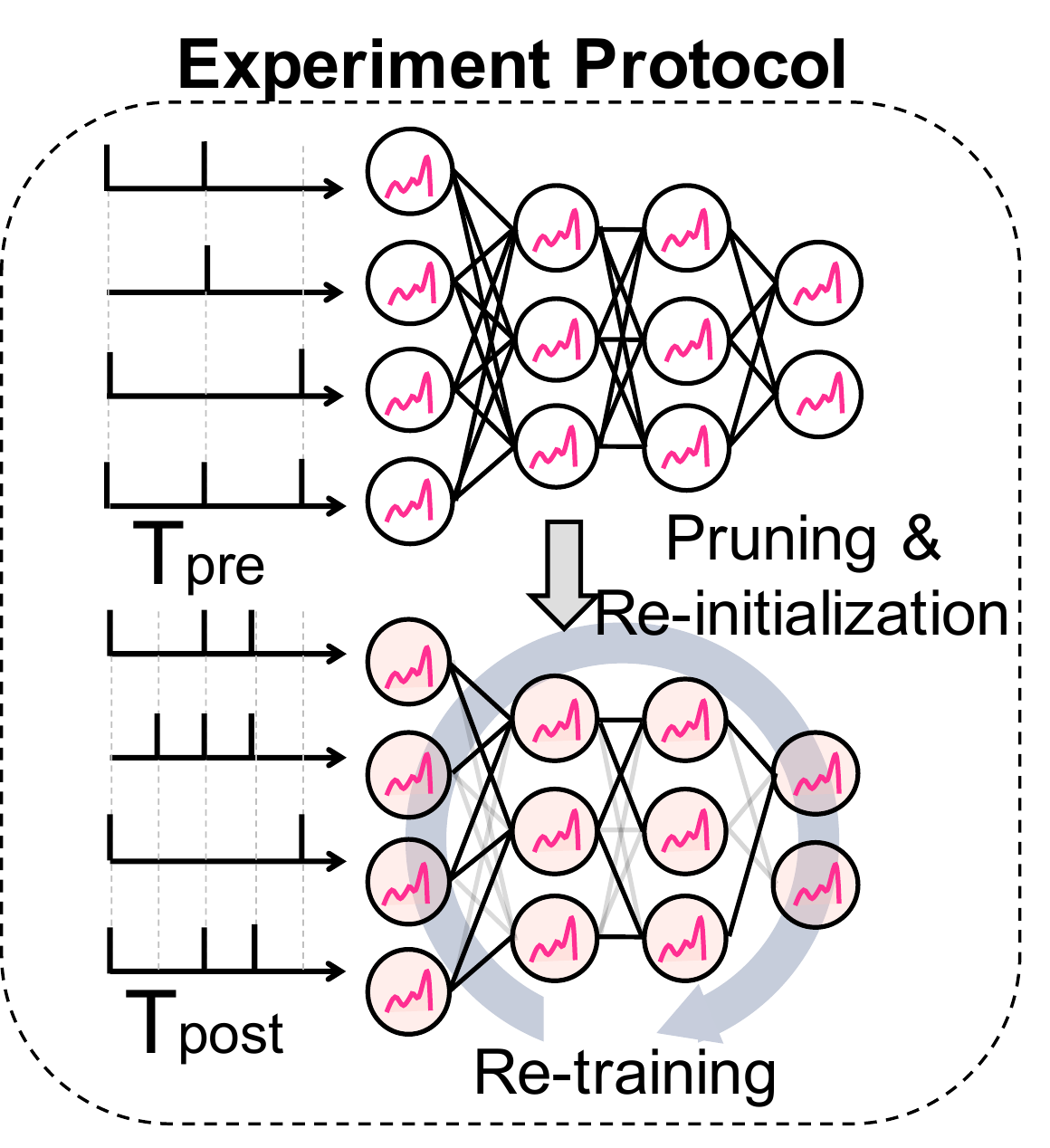} &
\includegraphics[width=0.215\linewidth]{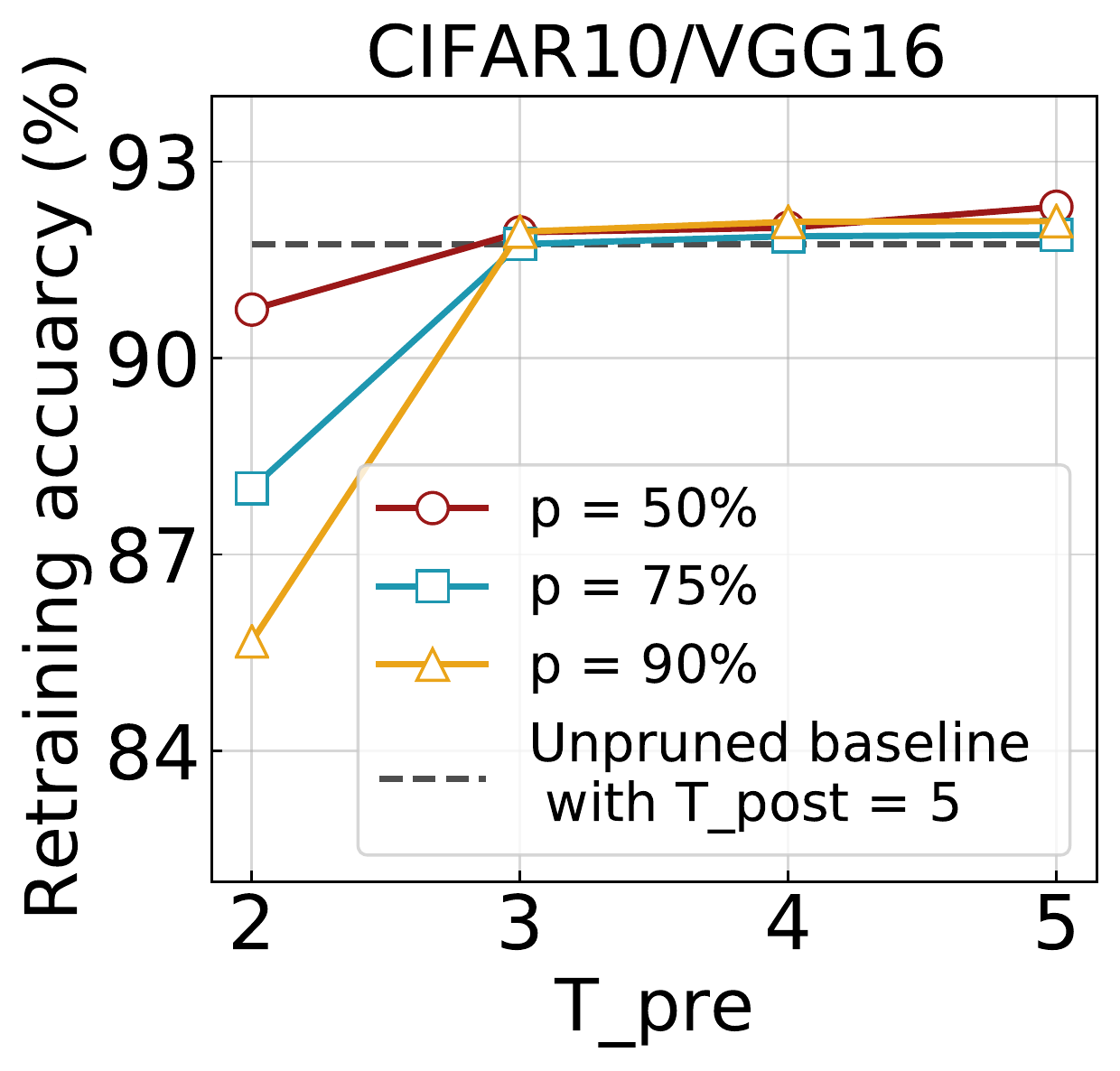} &
\includegraphics[width=0.197\linewidth]{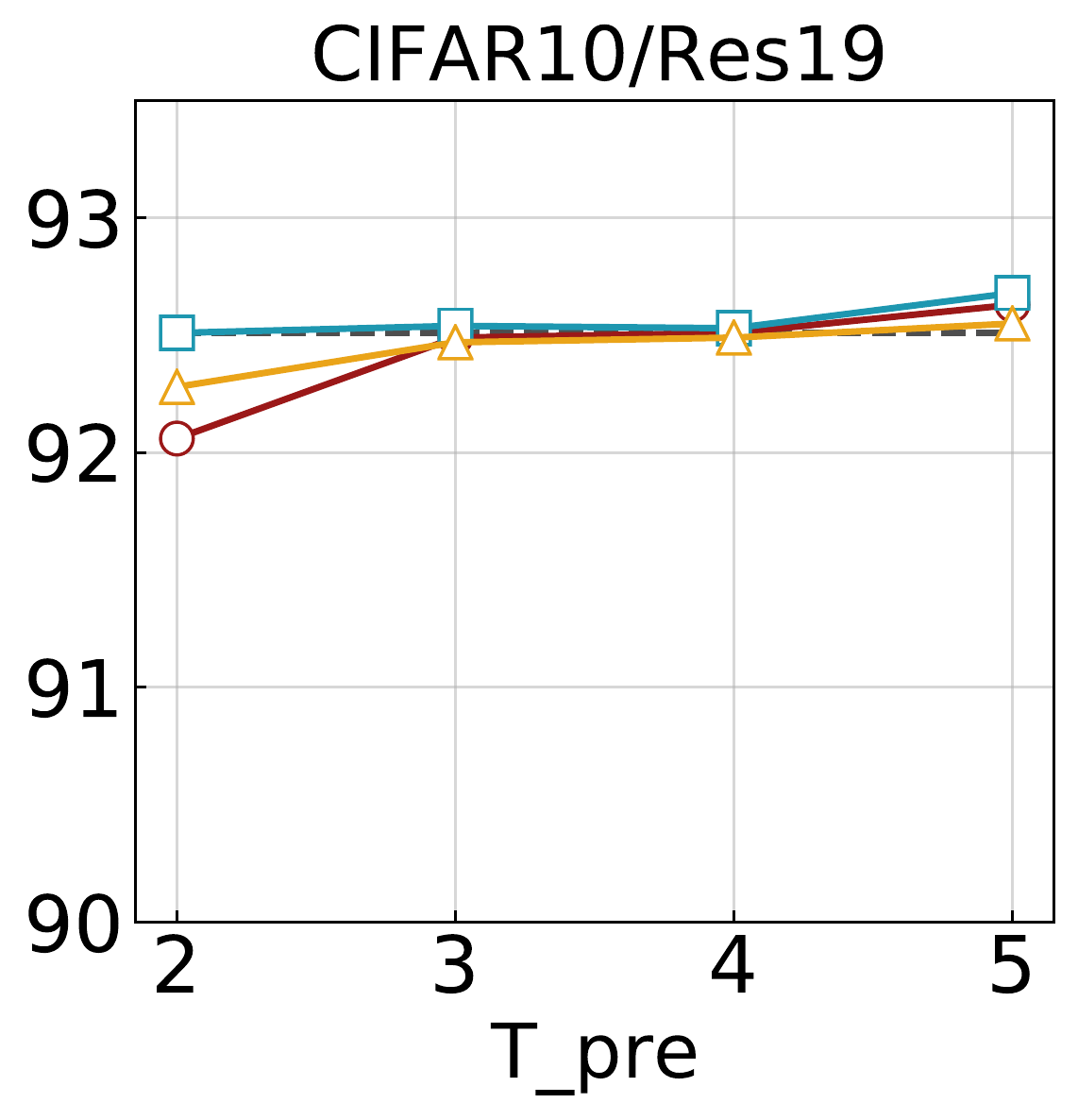} &
\includegraphics[width=0.197\linewidth]{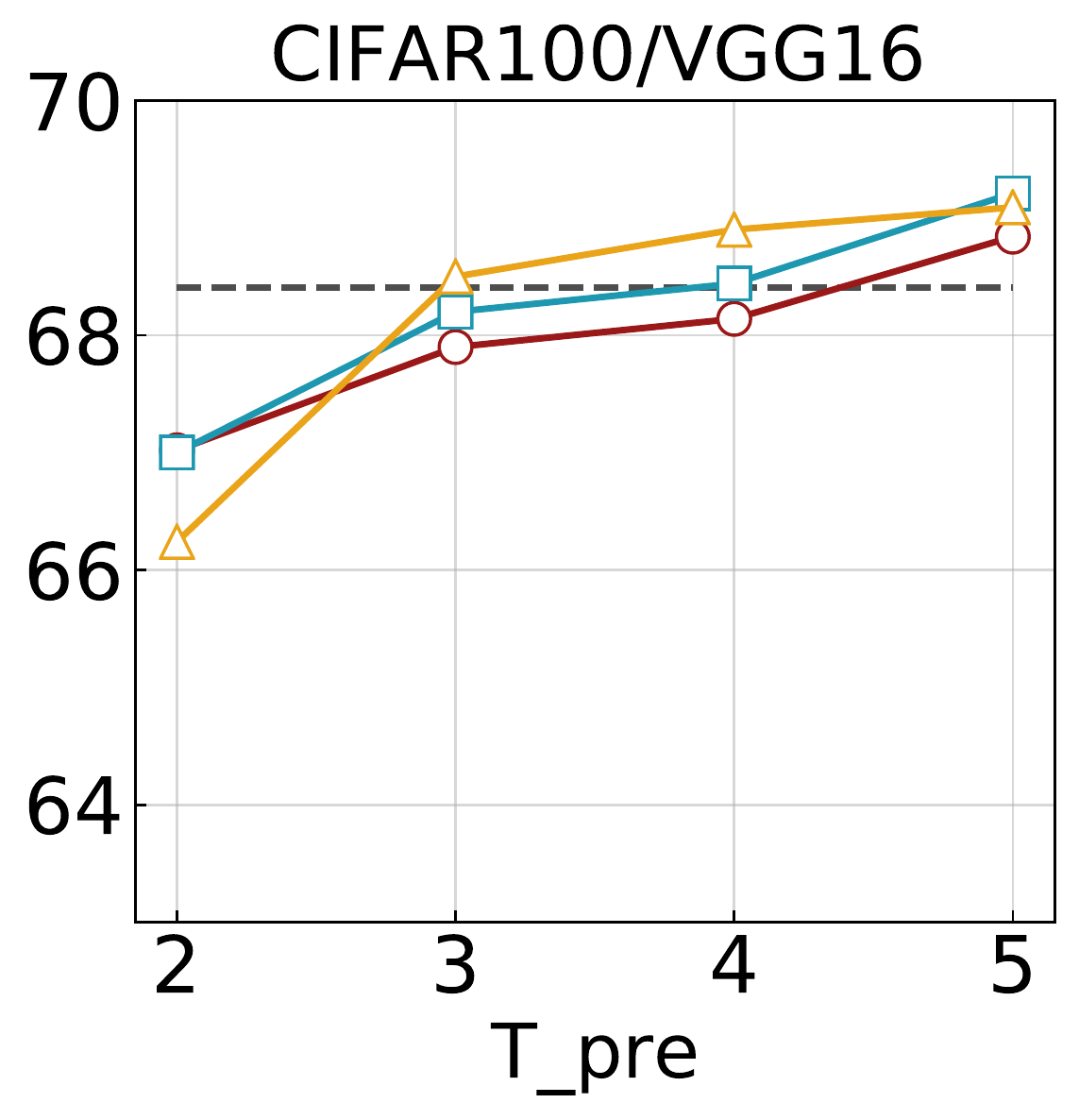}
 &
\includegraphics[width=0.197\linewidth]{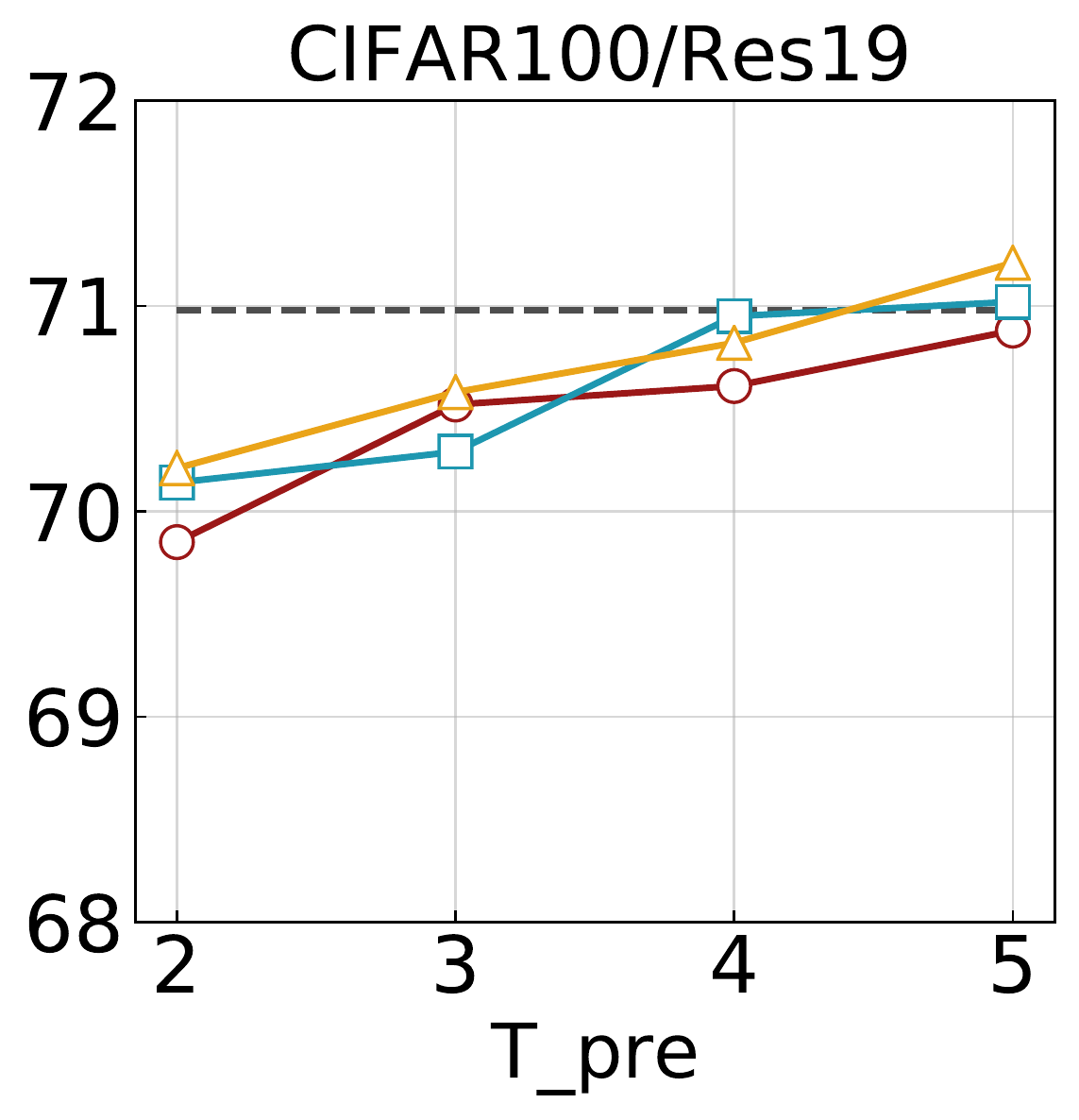}
\\
\end{tabular}
\caption{Preliminary experiments for Early-Time ticket.
We conduct experiments on VGG16/ResNet19 on CIFAR10/CIFAR100. We report retraining accuracy ($T_{post}=5$) with respect to the timestep ($T_{pre}$) for searching the important connectivity in SNNs.
}
\end{center}
\label{fig:method:timeticket_exp}
\end{figure}

\noindent{\textbf{Preliminary Experiments.} }
To answer this question, we conduct experiments on two representative deep architectures (VGG16 and ResNet19) on two datasets (CIFAR10 and CIFAR100).
Our experiment protocol is shown in Fig. \ref{fig:method:timeticket_exp} (left panel).
We first train the networks with timestep $T_{pre}$ till convergence.
After that,  we prune $p\%$ of the low-magnitude weights
and re-initialize the networks.
Finally, we re-train the pruned networks with longer timestep $T_{post} > T_{pre}$, and measure the test accuracy.
Thus, this experiment shows the performance of SNNs where the structure is obtained from the lower timestep.
In our experiments, we set $T_{pre} = \{2, 3, 4, 5\}$ and $T_{post} = 5$.
Surprisingly, the connections founded from $T_{pre} \ge 3$ 
can bring similar and even better accuracy compared to the unpruned baseline, as shown in Fig. \ref{fig:method:timeticket_exp}.
Note, in the preliminary experiments, we use a common post-training pruning based on the magnitude of weights \cite{han2015learning}.
Thus, we can extrapolate the existence of early-time winning tickets  
to a more sophisticated pruning method such as IMP \cite{frankle2018lottery}, which generally shows better performance than post-training pruning.
We call such a winning ticket as \textit{Early-Time tickets}; a winning ticket  drawn with a trained network from a smaller number of timesteps $T_{early}$, which shows matching performance with a winning ticket from the original timesteps $T$.

\begin{figure}[t]
\begin{center}
\def\arraystretch{0.5}
\begin{tabular}{@{}c@{}c@{}c@{}c@{}c@{}c}
\includegraphics[width=0.25\linewidth]{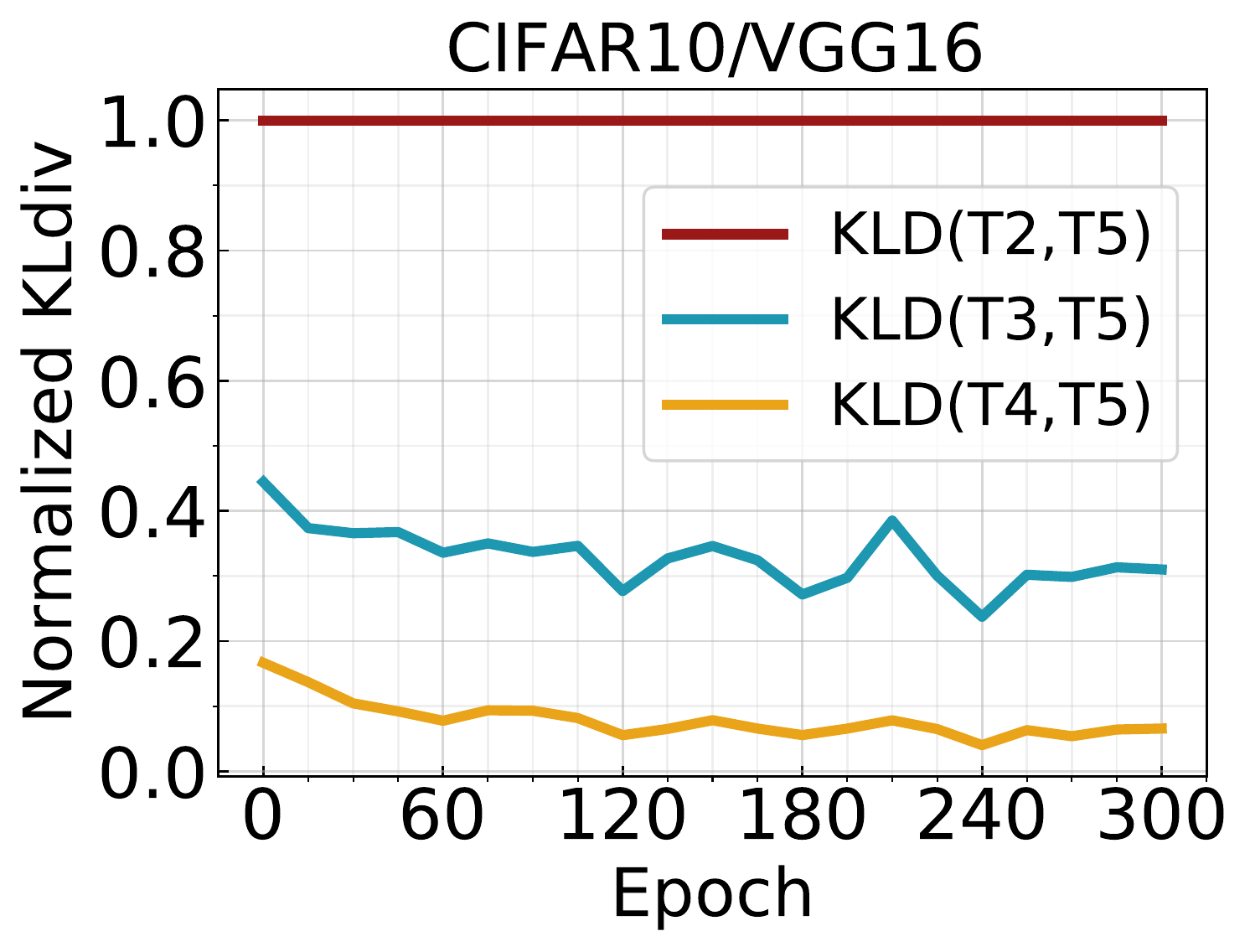} &
\includegraphics[width=0.25\linewidth]{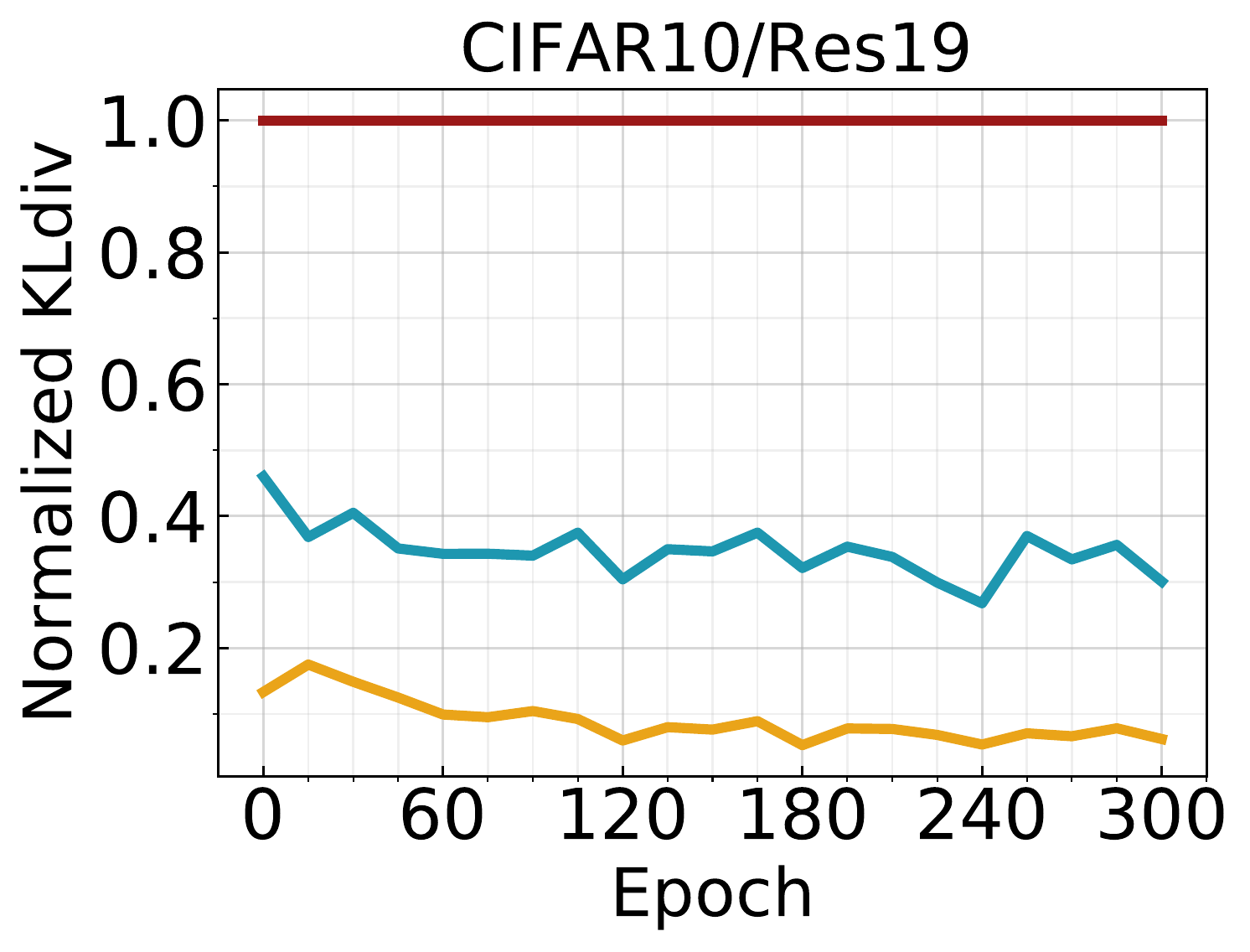} &
\includegraphics[width=0.25\linewidth]{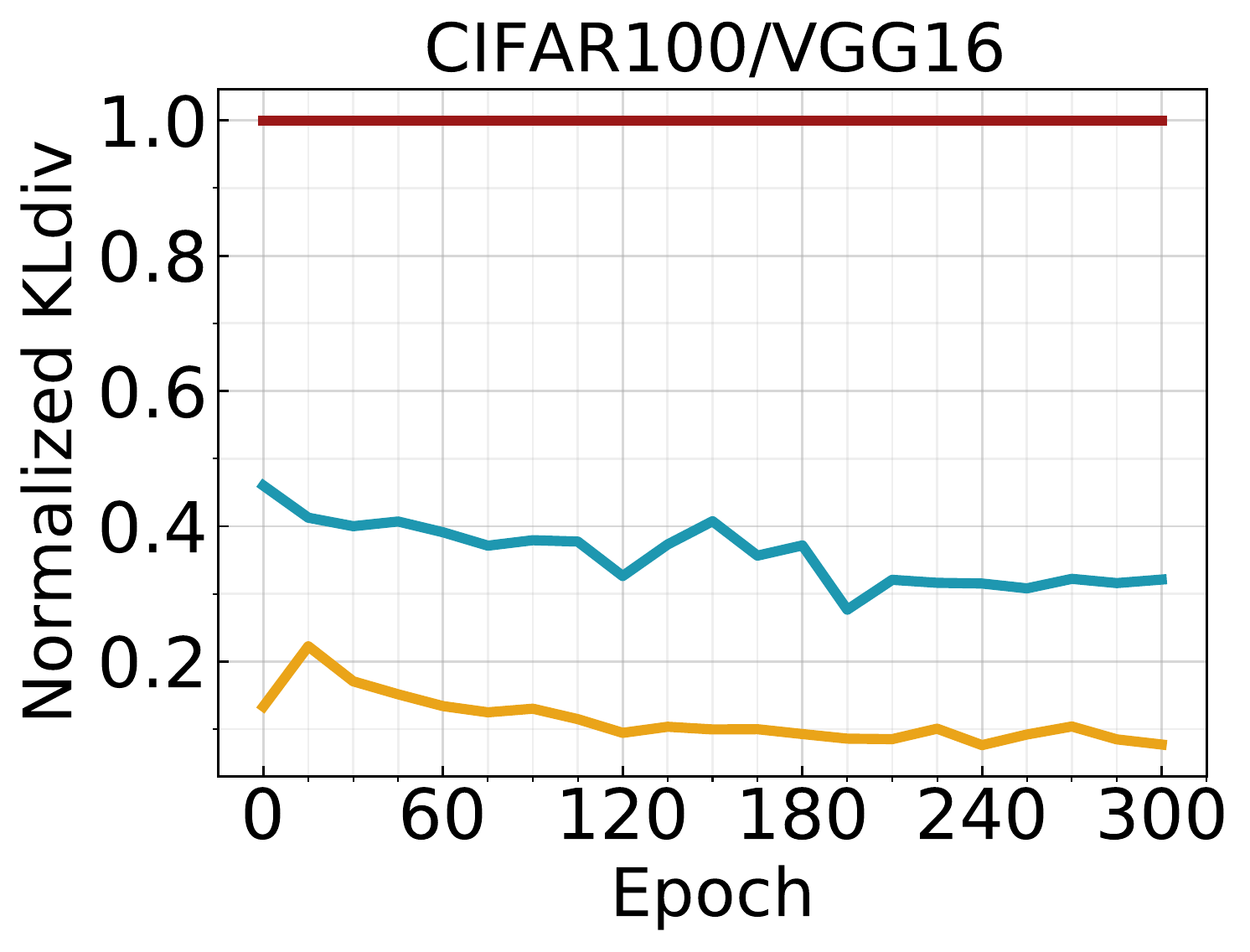}
 &
\includegraphics[width=0.25\linewidth]{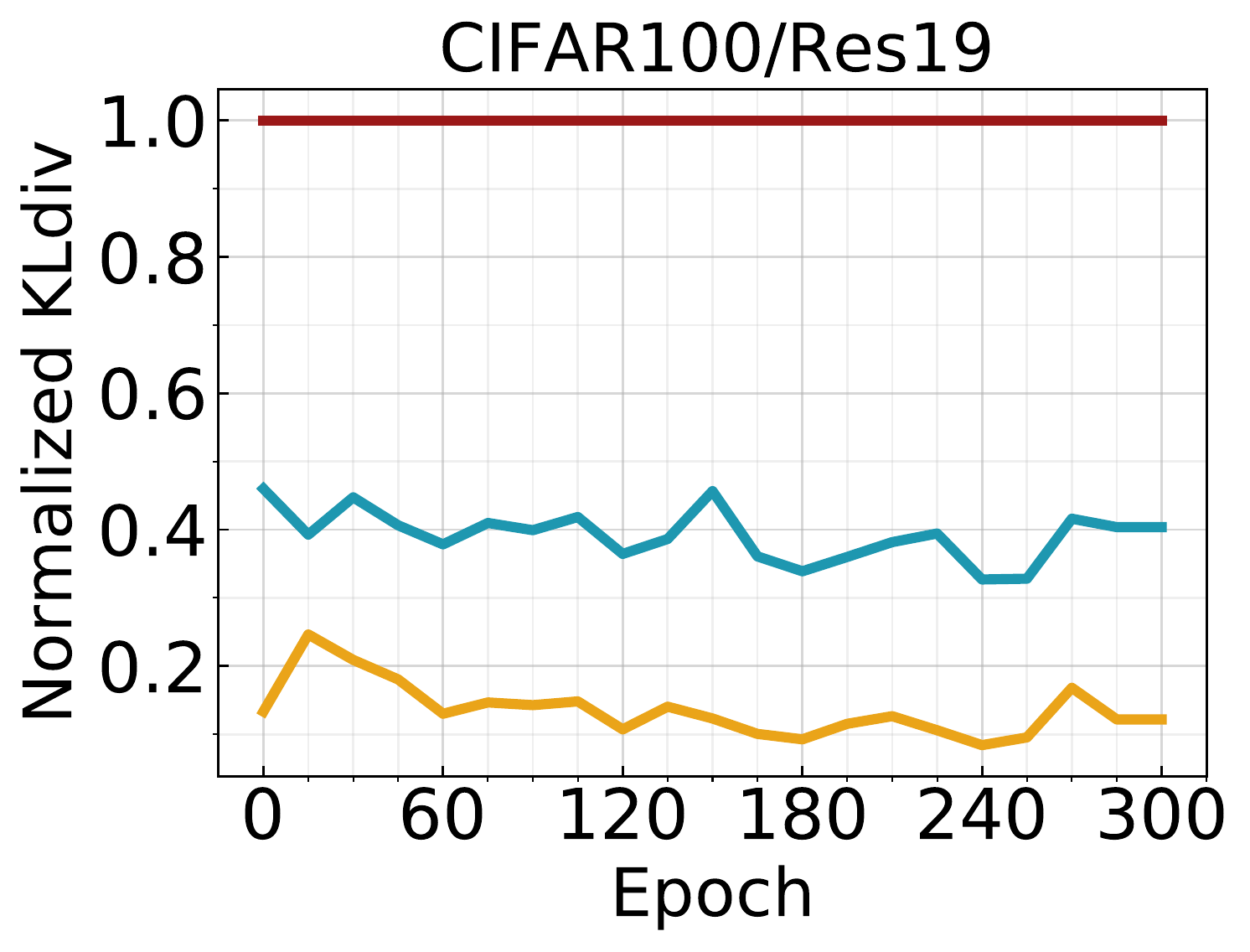}
\\
\end{tabular}
\caption{Kullback–Leibler (KL) divergence between the class prediction distribution from different timesteps.
The network is trained with the original timestep $T=5$.
We measure KL divergence between the predicted class probabilities from different timesteps.
We use the training set for calculating KL divergence.
}
\end{center}
\label{fig:method:kldiv}
\end{figure}

\begin{algorithm}[t]\small
        \caption{Early-Time (ET) ticket}
       \textbf{Input}: Training data $D$; Winning ticket searching method $F(\cdot)$ -- IMP or EB ticket; Original timestep $T$;   Threshold $\lambda$ \\
      \textbf{Output}:  Pruned  $SNN_{pruned}$
      \begin{algorithmic}[1]
        %
        \State {Training $SNN$ with N epochs for stability}
        \State{Memory = [ ]}
        \For{$t \gets 2$ to $T$}
        \State { $P_t \gets$ $SNN$($t$, $D$)} \Comment{Storing class prediction from each timestep}
        \State { Memory.append($P_t$)}
        \EndFor
        \State { $ [D_{KL}(P_{{T-1}}||P_{T}),...,D_{KL}(P_{{2}}||P_{T})] \gets$ Memory}
        \Comment{Computing KL div.}
        \For{$t \gets 2$ to $T-1$}
        \State {$\hat{D}_{KL}(P_{{t}}||P_{T}) \gets \frac{D_{KL}(P_{{t}}||P_{T})}{D_{KL}(P_{{2}}||P_{T})}$} \Comment{Normalization}
        \If{$\hat{D}_{KL}(P_{{t}}||P_{T})< \lambda$} \Comment{Select timestep when KL div. is less than $\lambda$}
                \State {$T_{early} = t$}
                \State {\textbf{break}}
        \EndIf
        \EndFor
       \State {$WinningTicket \gets F(SNN, T_{early})$} \Comment{Finding winning ticket with $T_{early}$}
       \State {$SNN_{pruned} \gets F(WinningTicket, T)$} \Comment{Train with the original timestep $T$}
       \State {\textbf{return} $SNN_{pruned}$}
      \end{algorithmic}
          \label{algorithm: overall}
\end{algorithm}

\noindent{\textbf{Proposed Method.} }
Then, \textit{how to practically select a timestep for finding  Early-Time tickets?}
The main idea is to measure the similarity between class predictions between the original timestep $T$ and a smaller number of timesteps, and select a minimal timestep that shows a similar representation with the target timestep.
Specifically, let $P_{T}$ be the class probability from the last layer of networks by accumulating output values across $T$ timesteps  \cite{lee2020enabling}. 
In this case, our search space can be $S = \{2, 3, ..., T-1\}$. Note, timestep 1 is not considered since it cannot use the temporal behavior of LIF neurons.
To measure the statistical distance between class predictions $P_{T}$ and $P_{T'}$, we use Kullback–Leibler (KL) divergence:
\begin{equation}
        D_{KL}(P_{{T'}}||P_{T}) = \sum_{x} P_{{T'}}(x) \textup{ln} \frac{P_{{T'}}(x)}{P_{{T}}(x)}.
\end{equation}
The value of KL divergence goes smaller when the timestep $T'$ is closer to the original timestep $T$, \ie $D_{KL}(P_{{T-1}}||P_{T})\le D_{KL}(P_{{T-2}}||P_{T})\le...\le D_{KL}(P_{{2}}||P_{T})$.
Note, for any $t\in \{1,..,T\}$, we compute $P_t$ by accumulating output layer's activation from $1$ to $t$ timesteps.
Therefore, due to the accumulation, if the timestep difference between timestep $t$ and $t'$ becomes smaller, the KL divergence $D_{KL}(P_{{t'}}||P_{t})$ becomes lower. 
After that, we rescale all KL divergence values to [0, 1] by dividing them with $D_{KL}(P_{{2}}||P_{T})$.
In Fig. \ref{fig:method:kldiv}, we illustrate normalized KL divergence of VGG16 and ResNet19 on CIFAR10 and CIFAR100, where we found two observations:
(1) 
The KL divergence with $T=2$ has a relatively higher value than other timesteps.
If the difference in the class probability is large (\ie larger KL divergence), weight connections are likely to be updated in a different direction.
This supports our previous observation that important connectivity founded by $T=2$ shows huge performance drop at $T=5$ (Fig. \ref{fig:method:timeticket_exp}).
Therefore, we search $T_{early}$ that has KL divergence less than $\lambda$ while minimizing the number of timesteps.
The $\lambda$ is a hyperparameter for the determination of $T_{early}$ in the winning ticket search process (Algorithm 1, line 15).
A higher $\lambda$ leads to smaller $T_{early}$ and vice versa (we visualize the change of $T_{early}$ versus different $\lambda$ values in Fig. \ref{fig:exp:property_EB_ET}(b)). 
For example, if we set $\lambda = 0.6$, timestep 3 is used for finding early-time tickets. 
In our experiments, we found that a similar value of threshold $\lambda$ can be applied across various datasets.
(2) The normalized KL divergence shows fairly consistent values across training epochs. Thus, we can find the suitable timestep $T_{early}$ for obtaining early-time tickets at the very beginning of the training phase.

The early-time ticket approach can be seamlessly applied to both IMP and Early-bird ticket methods.
Algorithm 1 illustrates the overall process of Early-Time ticket.
For stability with respect to random initialization, we start to search $T_{early}$ after $N=2$ epoch of training (Line 1). We show the variation of KL divergence results across training epochs in Supplementary C. 
We first find the $T_{early}$ from KL Divergence of difference timesteps (Line 2-13). 
After that, we discover the winning ticket using either IMP or EB ticket with $T_{early}$ (Line 14).
Finally, the winning ticket is trained with the original timestep $T$ (Line 15).

\begin{figure} [t]
    \centering
     \includegraphics[width=0.23\linewidth]{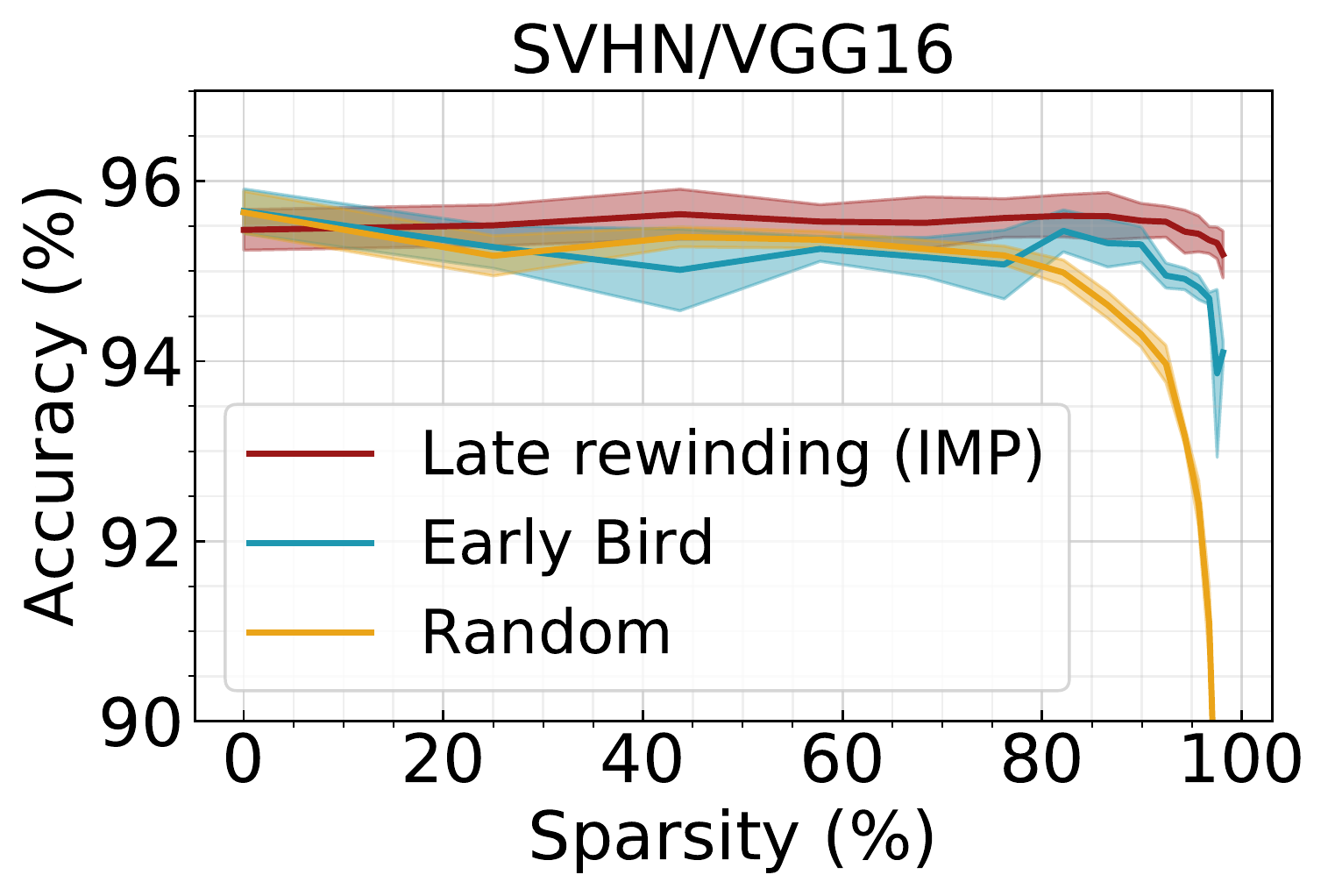} 
     \includegraphics[width=0.23\linewidth]{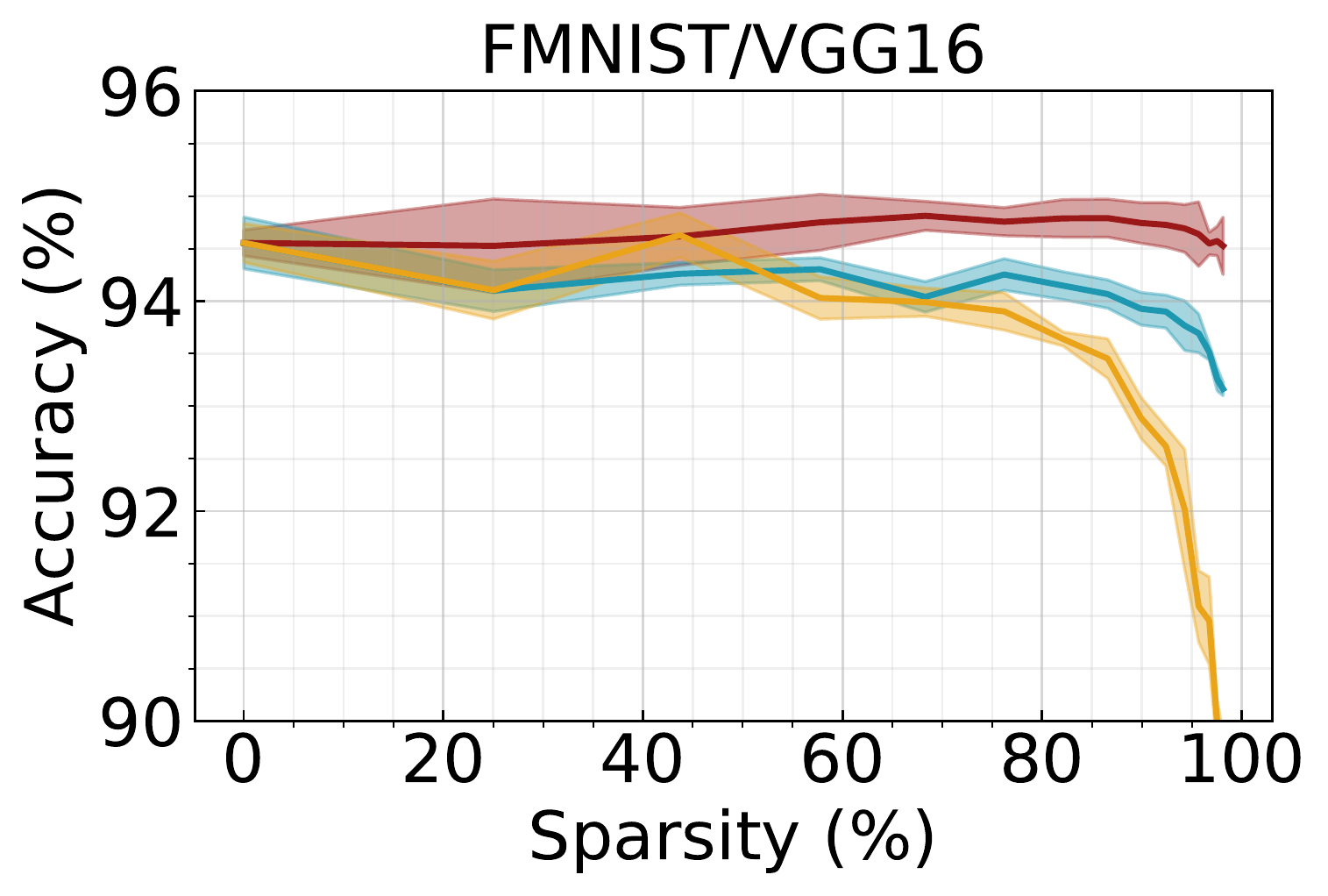}
    \includegraphics[width=0.23\linewidth]{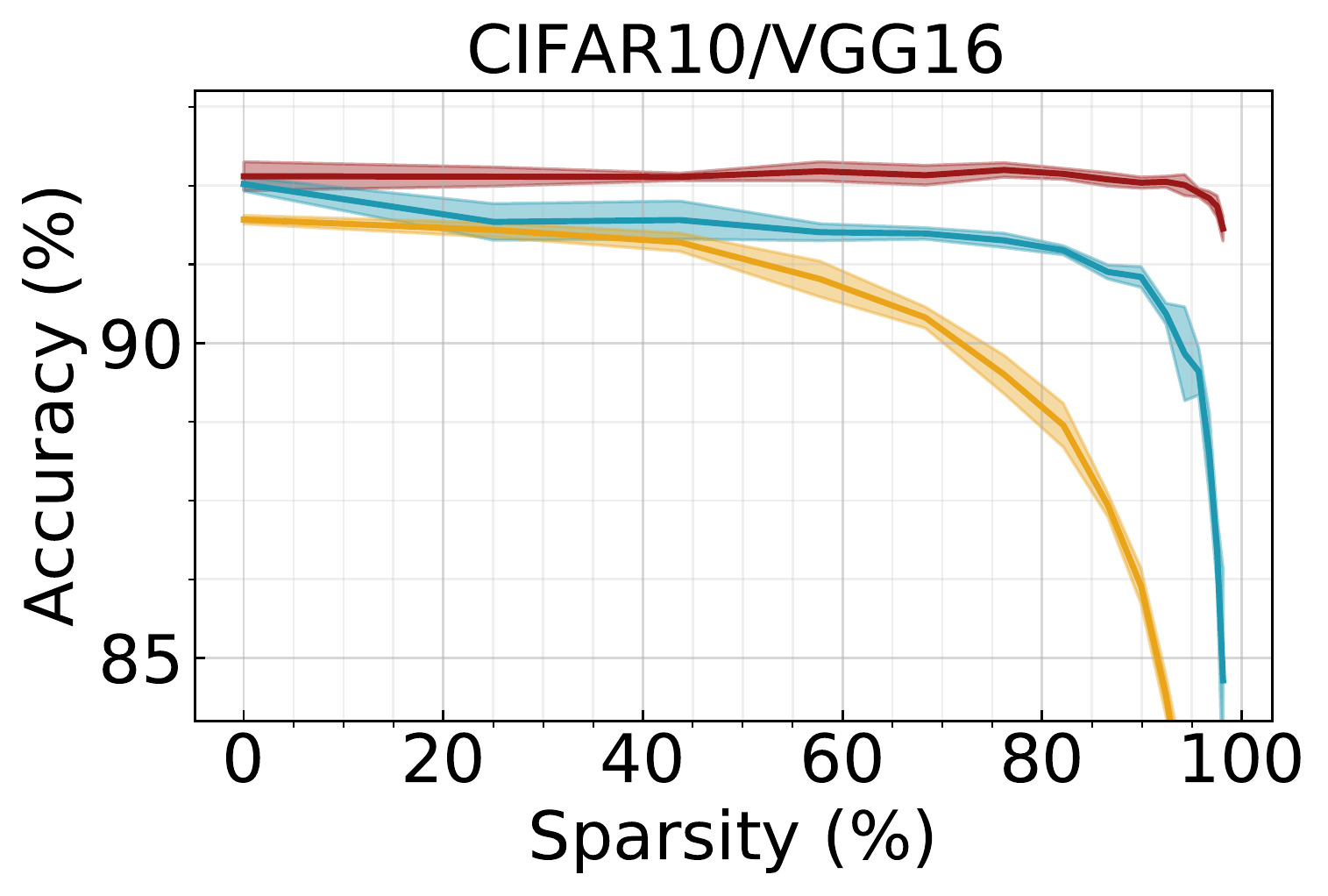}
 \includegraphics[width=0.23\linewidth]{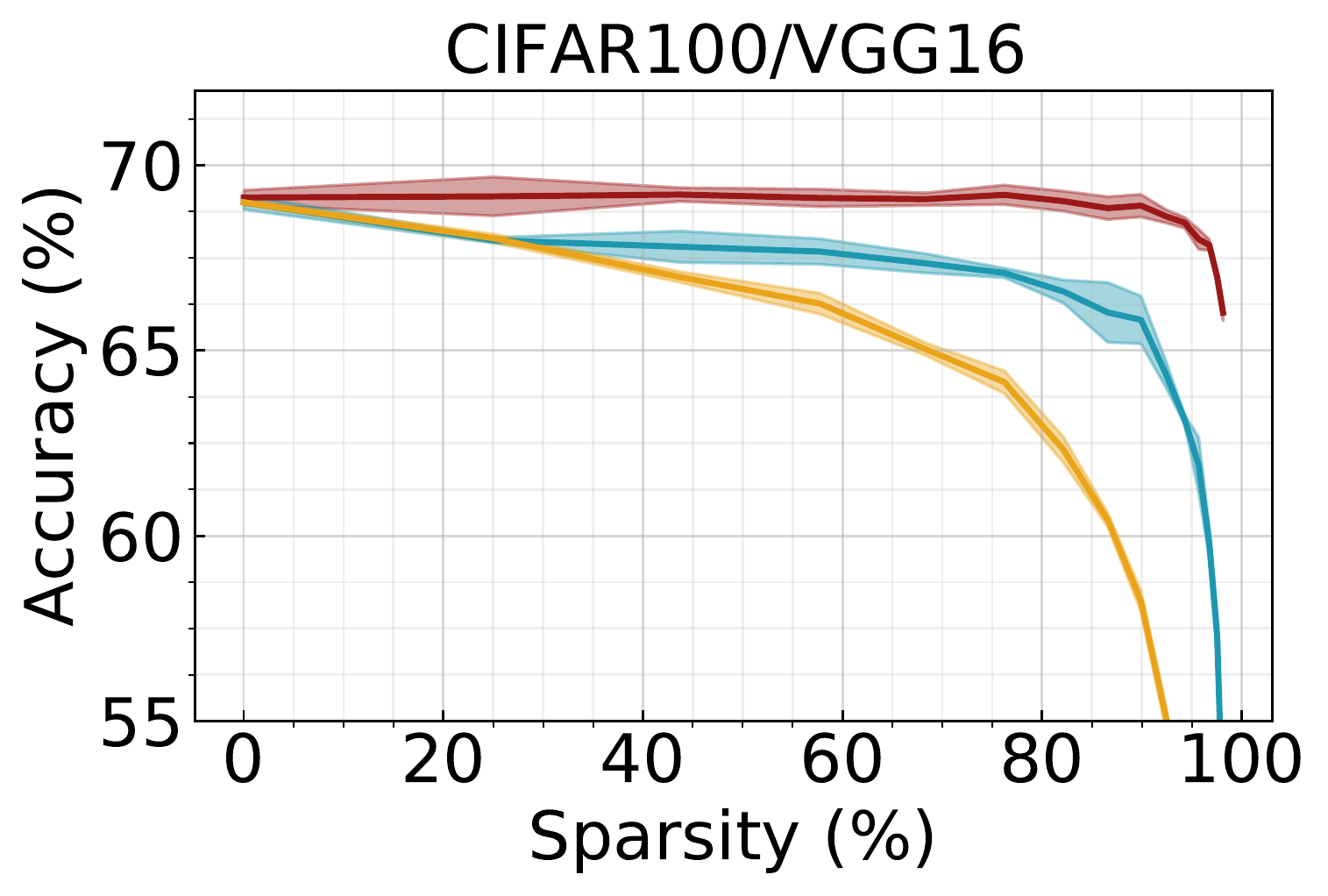} 
    \\
    \includegraphics[width=0.23\linewidth]{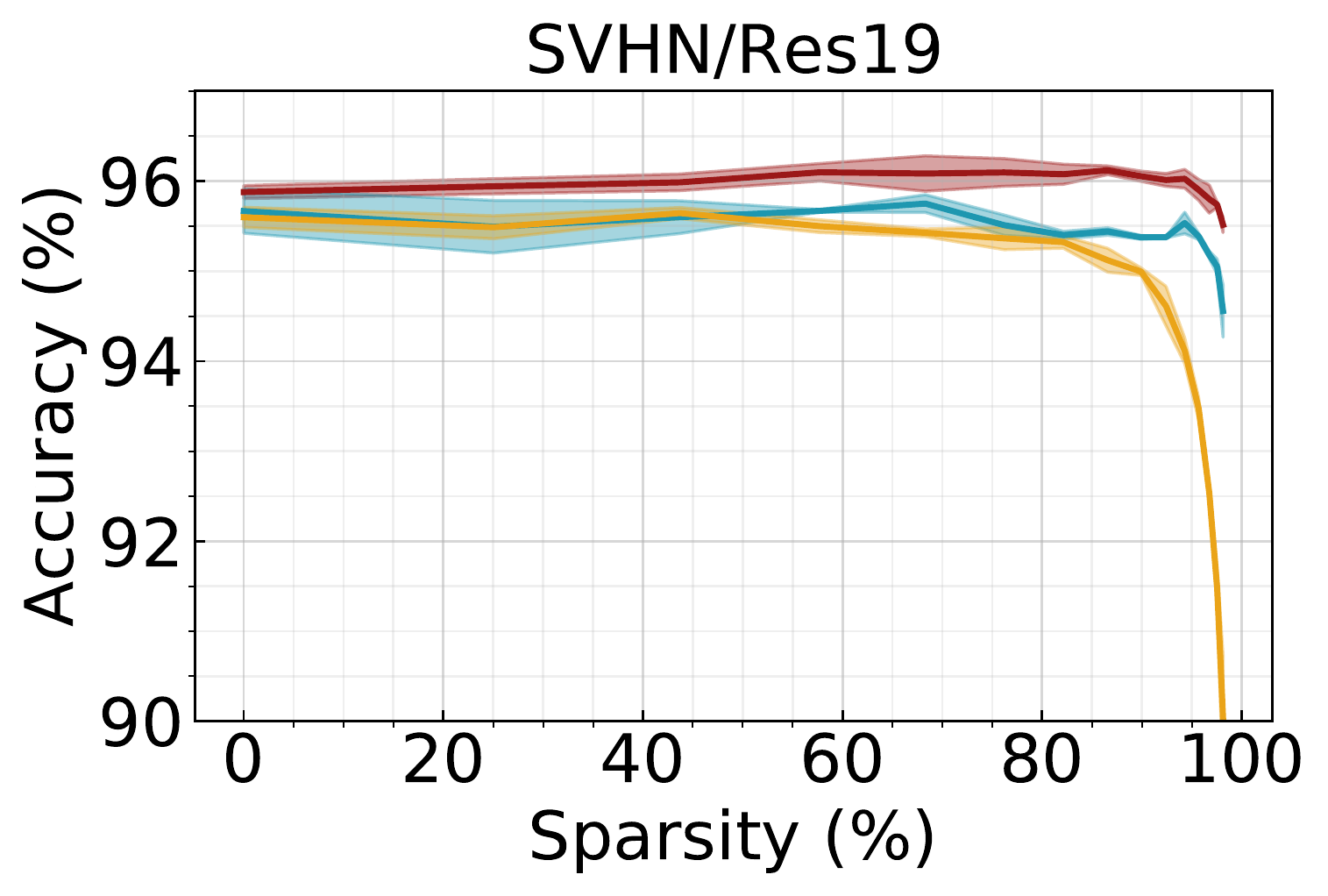} 
    \includegraphics[width=0.23\linewidth]{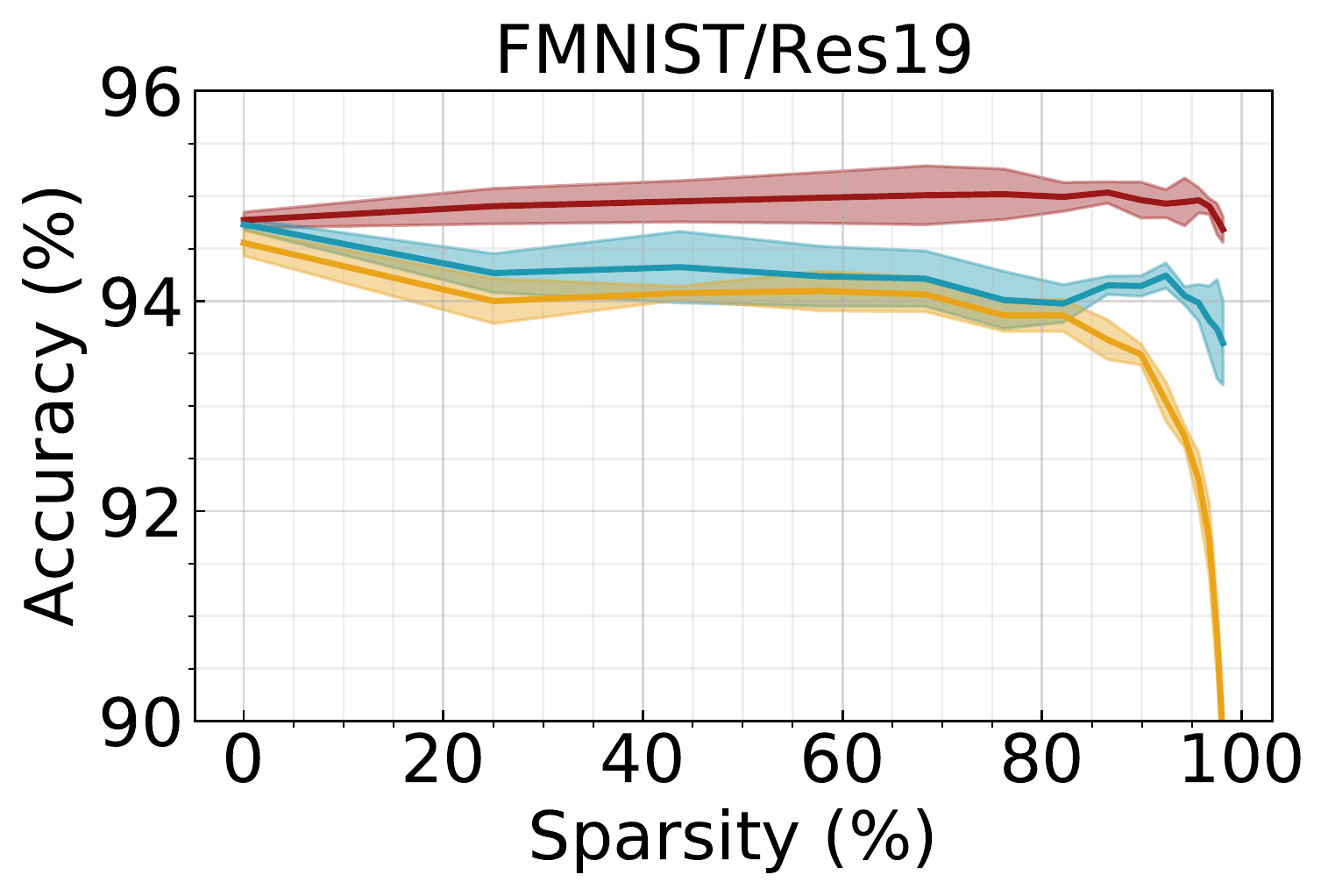} 
    \includegraphics[width=0.23\linewidth]{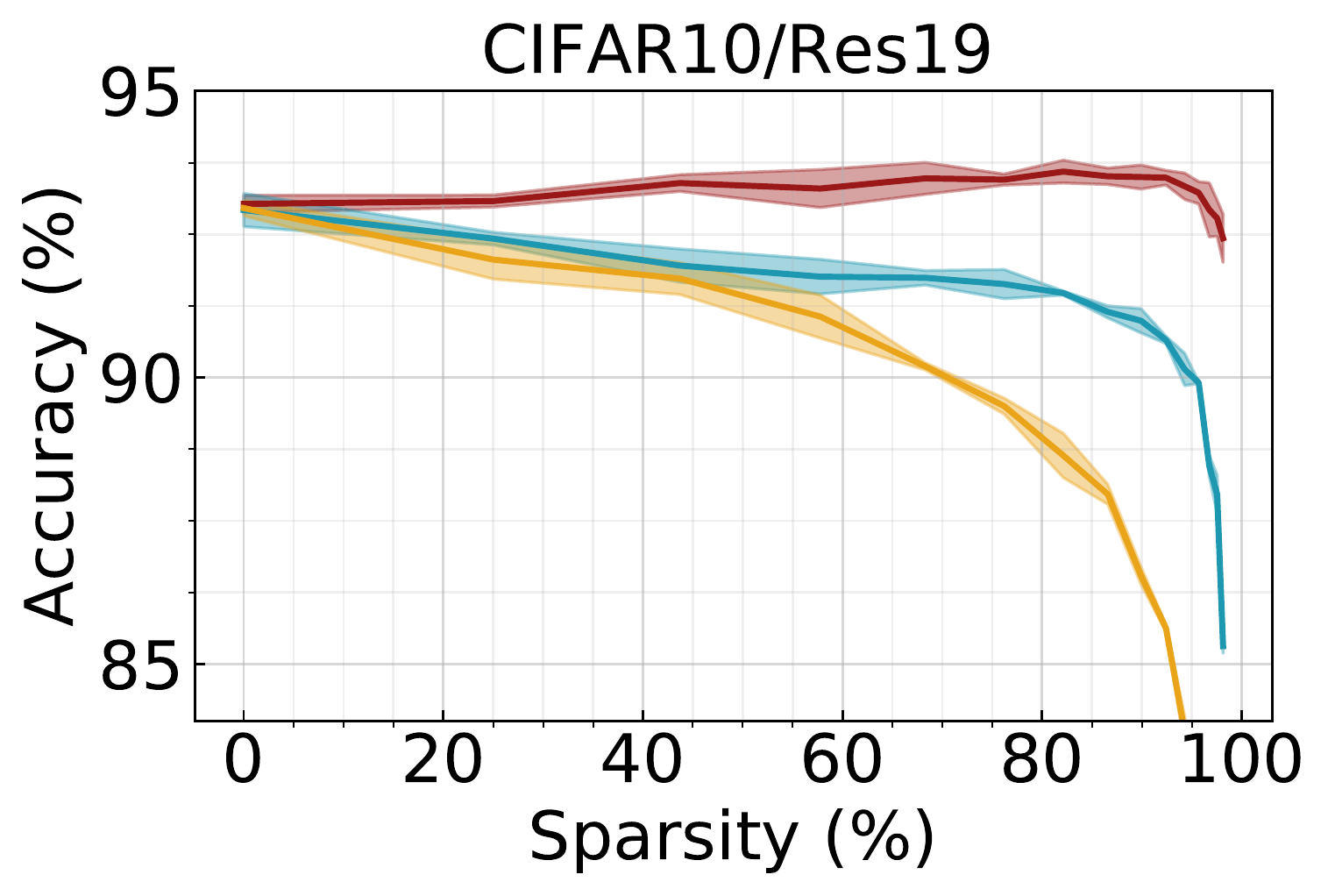} 
    \includegraphics[width=0.23\linewidth]{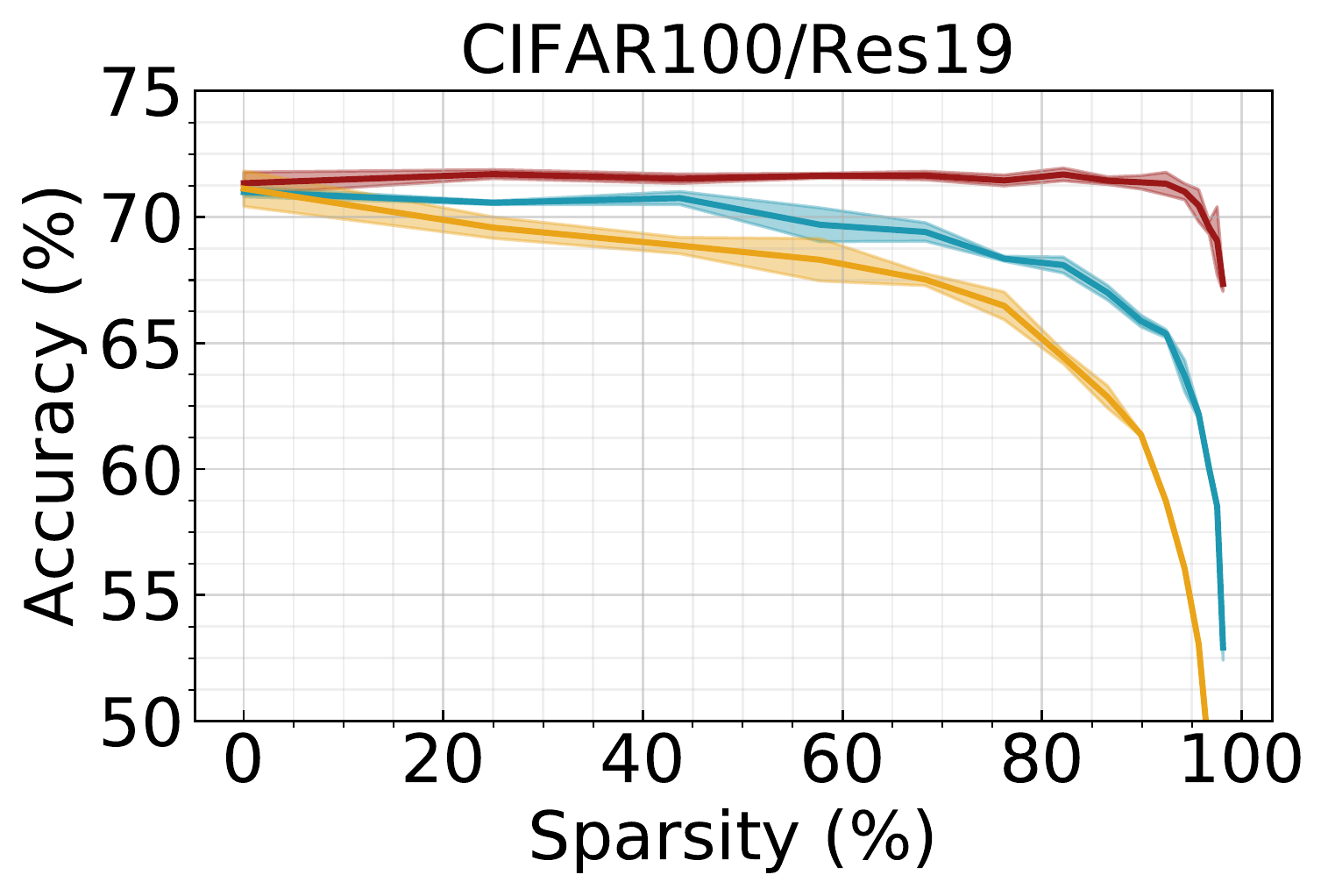}
    \caption{The accuracy of winning ticket with respect to sparsity level. We report mean and standard  deviation from 5 random runs.}
    \label{fig:exp:lth_early_random}
\end{figure}

\begin{table}[t!]
\addtolength{\tabcolsep}{0.5pt}
\centering
\caption{ Effect of the proposed Early-Time ticket. We compare the accuracy and search time of Iterative Magnitude Pruning (IMP), Early-Bird (EB) ticket, Early-Time (ET) ticket on four sparsity levels. We show search speed gain and accuracy change from applying ET. 
}
\resizebox{1.0\textwidth}{!}{%
\begin{tabular}{cl|cccc|cccc}
\toprule
\multirow{2}{*}{{Setting}}  & \multirow{2}{*}{{Method}\:}   & &\multicolumn{2}{c}{{Accuracy}}&& \multicolumn{4}{c}{{Winning Ticket Search Time (hours)}} \\
 \cline{3-10}&
 &\: p=68.30\%\:& \:p=89.91\% \:& \:p=95.69\%\:&\: p=98.13\%\:
&\: p=68.30\%\:& \:p=89.91\% \:& \:p=95.69\%\:&\: p=98.13\%\:  \\
 \cline{1-10}
 &  IMP  &  92.66 & 92.54 & 92.38  & 91.81  &  14.97 & 29.86 & 40.84 & 51.99\\
                   &  IMP + ET   &  92.49  &  92.09  &91.54 & 91.10 & 11.19  &	22.00  & 30.11  &	38.26  \\
                                       \cline{2-10}
             CIFAR10       & \cellcolor{black!5} $\Delta$ Acc. /  Speed Gain& \cellcolor{black!5} -0.17  &\cellcolor{black!5}   -0.45 &\cellcolor{black!5} -0.84& \cellcolor{black!5} -0.71&\cellcolor{black!5}  $\times1.34$ &	\cellcolor{black!5} $\times1.35$ &\cellcolor{black!5} $\times1.35$ &\cellcolor{black!5} $\times1.35$\\
                    \cline{2-10}
                  VGG16 &  EB    & 91.74 &   91.05 & 89.55  & 84.64 &
                  1.96	 & 0.74 &	0.11 &	0.09  \\
                  &  EB + ET    &  91.27   & 90.66  & 88.95& 84.86 
                  & 1.44  & 	0.55 &	0.07 &	0.06 \\
                                                         \cline{2-10}
                  &  \cellcolor{black!5} $\Delta$ Acc. /  Speed Gain & \cellcolor{black!5} -0.47   &\cellcolor{black!5}  -0.39 &\cellcolor{black!5} -0.60 &\cellcolor{black!5} +0.22
                  &\cellcolor{black!5} $\times1.36$ & \cellcolor{black!5}	$\times1.34$ &\cellcolor{black!5}	 $\times1.18$&\cellcolor{black!5}	$\times1.12$ \\
                  
\hline 
\hline 
  &  IMP  &  93.47 &  93.49 & 93.22 &  92.43 & 21.01& 42.20 &58.91 &73.54     \\
                   &  IMP + ET  &  93.10 & 92.72&  92.68    &91.36 &  13.35 & 26.62&37.27&46.40     \\ \cline{2-10}
              CIFAR10      & \cellcolor{black!5} $\Delta$ Acc. /  Speed Gain& -0.37 \cellcolor{black!5}  &-0.77 \cellcolor{black!5}   &-0.54\cellcolor{black!5} & -1.07 \cellcolor{black!5} &\cellcolor{black!5}  $\times$1.57 &	\cellcolor{black!5} $\times$1.59 &\cellcolor{black!5} $\times$1.58 &\cellcolor{black!5} $\times$1.58\\
                    \cline{2-10}
            Res19       &  EB     &  91.00 &90.84 & 89.90 & 85.22& 2.49   &0.87   &0.24&0.08   \\
                  &  EB + ET &  90.83 & 91.21& 89.65  & 85.45   &1.63   &0.58 &1.70 &0.07     \\\cline{2-10}
                    & \cellcolor{black!5} $\Delta$ Acc. / Speed Gain& \cellcolor{black!5}-0.17  &\cellcolor{black!5}+0.37   &\cellcolor{black!5}-0.50 & \cellcolor{black!5}-1.09 &\cellcolor{black!5}  $\times$1.52 &	\cellcolor{black!5} $\times$1.49 &\cellcolor{black!5} $\times$1.38 &\cellcolor{black!5} $\times$1.16\\
\hline 
\hline 
 &  IMP  &  69.08   & 68.90  & 68.00 &  66.02 & 15.02 & 29.99 & 41.03 & 52.05\\
  &  IMP + ET  &  68.27    &  67.99 & 66.51 & 64.41  & 11.24  &	22.42  & 30.53 & 38.32   \\
  \cline{2-10}
                CIFAR100    & \cellcolor{black!5} $\Delta$ Acc. /  Speed Gain& -0.81 \cellcolor{black!5}  &-0.91\cellcolor{black!5}   &-1.49\cellcolor{black!5} &-1.61 \cellcolor{black!5} &\cellcolor{black!5}  $\times$1.33&	\cellcolor{black!5} $\times$1.34 &\cellcolor{black!5} $\times$1.34 &\cellcolor{black!5} $\times$1.36\\
                    \cline{2-10}
              VGG16     &  EB &   67.35 &  65.82 &  61.90  & 52.11 &   2.27	& 0.99&0.32 &	0.06 \\
                  &  EB + ET  &  67.26  &   64.18   & 61.81 &52.77 &  1.66 & 	0.73 &	0.24&	0.05   \\
                  \cline{2-10}
                    & \cellcolor{black!5} $\Delta$ Acc. /  Speed Gain& \cellcolor{black!5}-0.09  &\cellcolor{black!5}-1.64   &\cellcolor{black!5}-0.09 & \cellcolor{black!5}+0.66 &\cellcolor{black!5}  $\times$1.36 &	\cellcolor{black!5} $\times$1.35 &\cellcolor{black!5} $\times$1.31 &\cellcolor{black!5} $\times$1.12\\
\hline 
\hline 
  &  IMP & 71.64 &   71.38   & 70.45&  67.35 & 21.21 &  42.29 &59.17 &73.52  \\
                   &  IMP + ET  &71.06  & 70.45 &   69.23   & 65.49& 13.56  & 27.05 & 37.88 & 46.65    \\\cline{2-10}
                   CIFAR100 & \cellcolor{black!5} $\Delta$ Acc. /  Speed Gain& -0.58 \cellcolor{black!5}  & -0.93 \cellcolor{black!5}   &-1.22\cellcolor{black!5} & -1.86\cellcolor{black!5} &\cellcolor{black!5}  $\times$1.56 &	\cellcolor{black!5} $\times$1.56 &\cellcolor{black!5} $\times$1.56 &\cellcolor{black!5} $\times$1.57\\
                    \cline{2-10} 
                 ResNet19  &  EB  & 69.41  & 65.87 & 62.18 & 52.92  & 3.08&    1.71&0.43 &0.09    \\
                  &  EB + ET&  68.98 &65.76 &  62.20 &   51.50  & 2.00  &1.12 & 0.29&0.07   \\
                  \cline{2-10}
                    & \cellcolor{black!5} $\Delta$ Acc. /  Speed Gain& \cellcolor{black!5} -0.43 &\cellcolor{black!5}  -0.12 &\cellcolor{black!5}+0.02& \cellcolor{black!5}-1.42 &\cellcolor{black!5}  $\times$1.53 &	\cellcolor{black!5} $\times$1.52 &\cellcolor{black!5} $\times$1.45 &\cellcolor{black!5} $\times$1.16\\
\bottomrule
\end{tabular}%
}
\label{table:exp:et_ticket}
\end{table}

\section{Experimental Results}

\subsection{Implementation Details}

We comprehensively evaluate various pruning methods on four public datasets: SVHN~\cite{netzer2011reading}, Fashion-MNIST~\cite{xiao2017fashion}, CIFAR10~\cite{krizhevsky2009learning}
and CIFAR100~\cite{krizhevsky2009learning}.
In our work, we focus on pruning deep SNNs, therefore we evaluate two representative architectures; VGG16 \cite{simonyan2014very} and ResNet19 \cite{he2016deep}.
Our implementation is based on PyTorch \cite{paszke2017automatic}.
We train the network using SGD optimizer with momentum 0.9 and weight decay 5e-4. Our image augmentation process and loss function follow the previous SNN work  \cite{deng2022temporal}.
We set the training batch size to 128. The base learning rate is set to 0.3 for all datasets with cosine learning rate scheduling~\cite{loshchilov2016sgdr}. Here, we set the total number of epochs to 150, 150, 300, 300 for SVHN, F-MNIST, CIFAR10, CIFAR100, respectively. 
We set the default timesteps $T$ to $5$ across all experiments. Also, we use $\lambda = 0.6$ for finding Early-Time tickets.
Experiments were conducted on an RTX 2080Ti GPU with PyTorch implementation. We use SpikingJelly \cite{SpikingJelly} package for implementation.

\subsection{Winning Tickets in SNNs}

\noindent\textbf{Performance of IMP and EB ticket.}   In Fig. \ref{fig:exp:lth_early_random}, we show the performance of the winning tickets from IMP and EB. The performance of random pruning is also provided as a reference.
Both IMP and EB can successfully find the winning ticket, which shows better performance than random pruning. 
Especially, IMP finds winning tickets over $\sim97\%$ sparsity across all configurations.
Also, we observe that the winning ticket sparsity is affected by dataset complexity.
EB ticket can find winning tickets ($>95\%$ sparsity) for relatively simple datasets such as SVHN and Fashion-MINST. However, they are limited to discovering the winning ticket having $\le95\%$ sparsity on CIFAR10 and CIFAR100.
We further provide experiments on ResNet34-TinyImageNet and AlexNet-CIFAR10 in Supplementary G.

\noindent\textbf{Effect of ET Ticket.} 
In Table \ref{table:exp:et_ticket}, we report the change in accuracy and search speed gain from applying ET to IMP and EB, on CIFAR10 and CIFAR100 datasets (SVHN and Fashion-MNIST results are provided in Supplementary E). 
Although IMP achieves highest accuracy across all sparsity levels, they require $26\sim73$ hours to search winning tickets with $98.13\%$ sparsity.
By applying ET to IMP, the search speed increases up to $\times 1.59$ without a huge accuracy drop (of course, there is an accuracy-computational cost trade-off because the ET winning ticket cannot exactly match with the IMP winning ticket).
Compared to IMP, EB ticket provides significantly less search cost for searching winning tickets.
Combining ET with EB ticket brings faster search speed, even finding a winning ticket in one hour (on GPU).
At high sparsity levels of EB ($p=98.13\%$), search speed gain from ET is upto $\times1.16$.
Also, applying ET on ResNet19 brings a better speed gain than VGG16 since ResNet19 requires a larger computational graph from multiple timestep operations (details are in Supplementary D). 
Overall, the results support our hypothesis that important weight connectivity of the SNN can be discovered from shorter timesteps.

\begin{figure}[t!]
\begin{center}
\def\arraystretch{0.5}
\begin{tabular}{@{}c@{}c@{}c}
\includegraphics[width=0.31\linewidth]{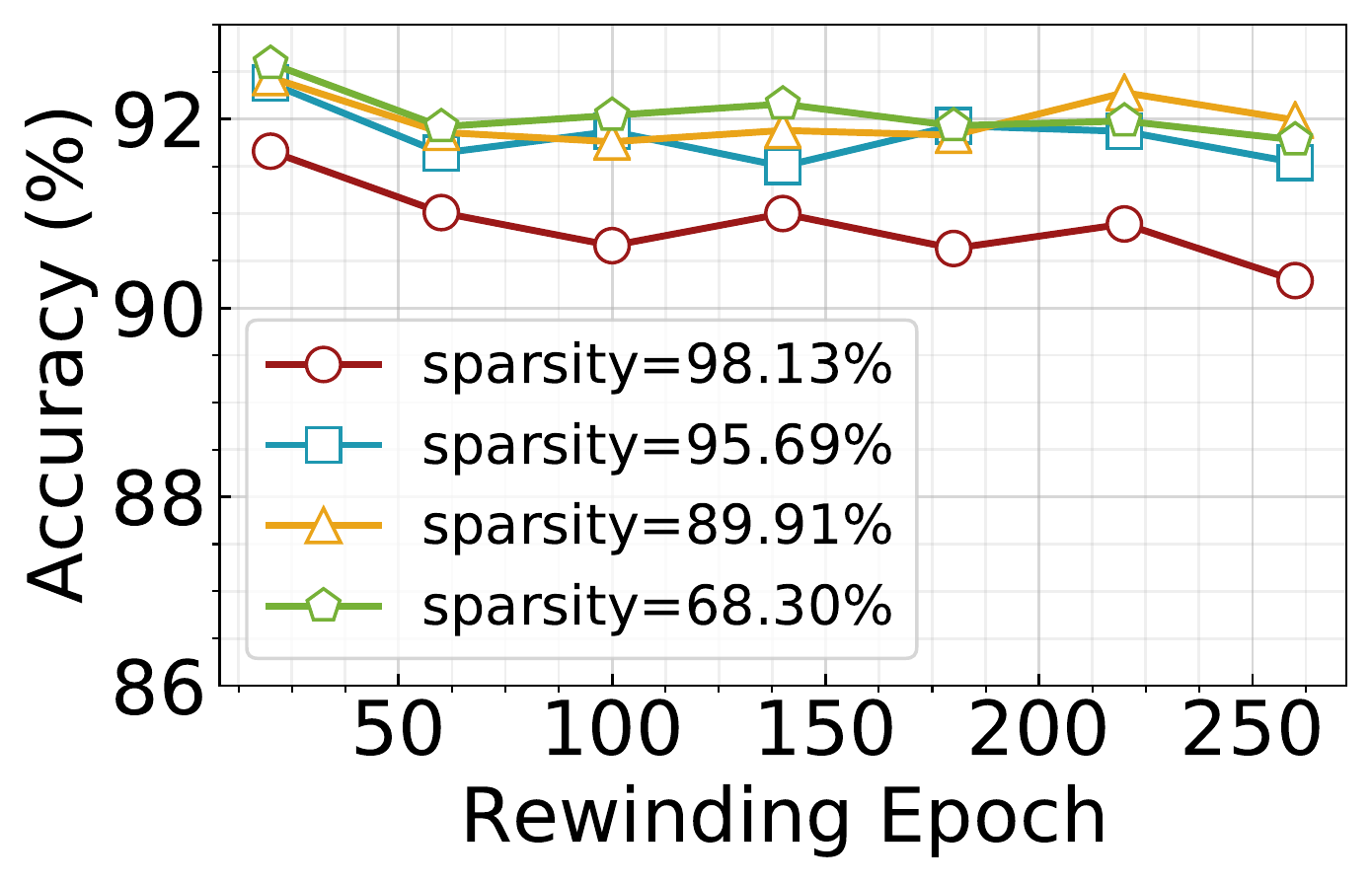} &
\includegraphics[width=0.31\linewidth]{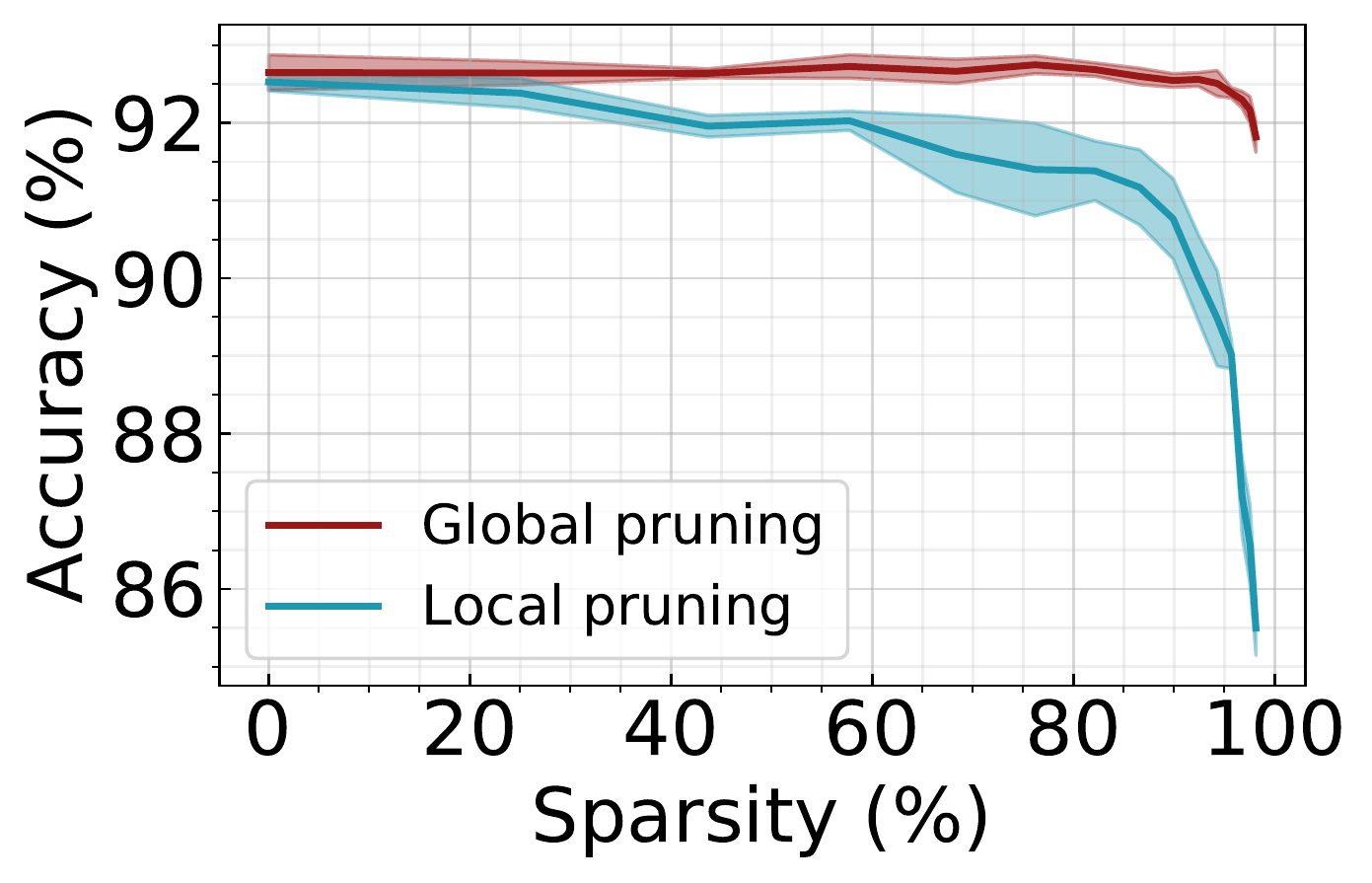}   &
\includegraphics[width=0.31\linewidth]{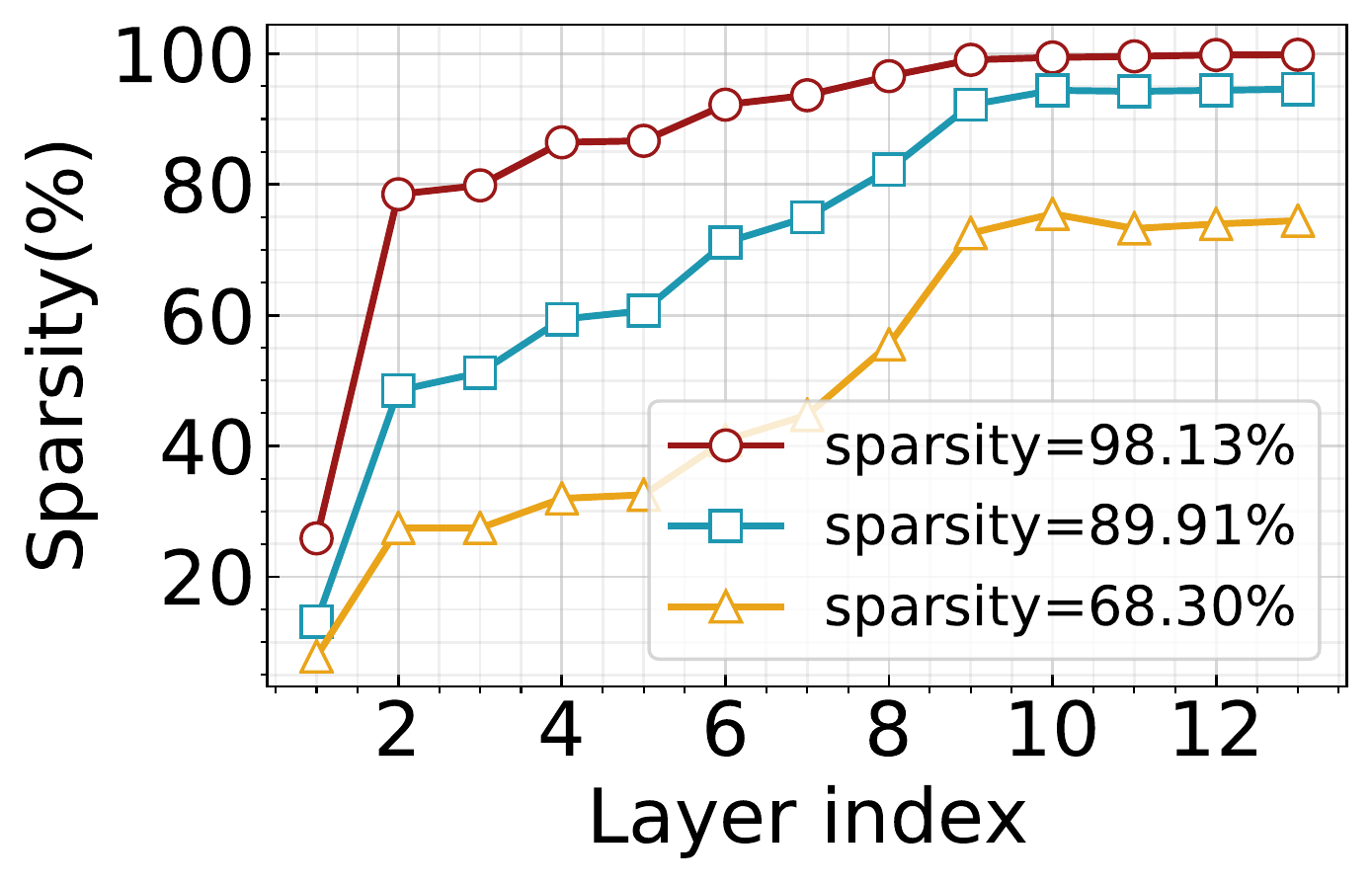}
\\
{ (a)  } & {(b) }& { (c) }\\
\end{tabular}
\caption{Observations from Iterative Magnitude Pruning (IMP). (a) The performance change with respect to the rewinding epoch. (b) The performance of global pruning and local pruning. (c) Layer-wise sparsity across different sparsity levels with global pruning. We use VGG16 on the CIFAR10 dataset for experiments.
}
\label{fig:exp:property_IMP}
\end{center}
\end{figure}

\noindent\textbf{Observations from Pruning Techniques.}  
We use \textit{Late Rewinding} \cite{frankle2019stabilizing} (IMP) for obtaining stable performance at high sparsity regime (refer Section 3.1).
To analyze the effect of the rewinding epoch, we change the rewinding epoch and report the accuracy on four sparsity levels in Fig. \ref{fig:exp:property_IMP}(a).
We observe that the rewinding epoch does not cause a huge accuracy change with sparsity $\le 95.69\%$.
However, a high sparsity level ($98.13\%$) shows non-trivial accuracy drop $\sim1.5\%$ at epoch $260$, which requires careful rewinding epoch selection.
Frankle \etal \cite{frankle2019stabilizing} also show that using the same pruning percentage for both shallow and deep layers (\ie local pruning) degrades the accuracy.
Instead of local pruning, they apply different pruning percentages for each layer (\ie global pruning).
We compare global pruning and local pruning in Fig. \ref{fig:exp:property_IMP}(b).
In SNN, global pruning achieves better performance than local pruning,  especially for high sparsity levels.
Fig. \ref{fig:exp:property_IMP}(c) illustrates layer-wise sparsity obtained from global pruning.
The results show that deep layers have higher sparsity compared to shallow layers.

\begin{figure}[t!]
\begin{center}
\def\arraystretch{0.5}
\begin{tabular}{@{}c@{}c@{}c}
\includegraphics[width=0.31\linewidth]{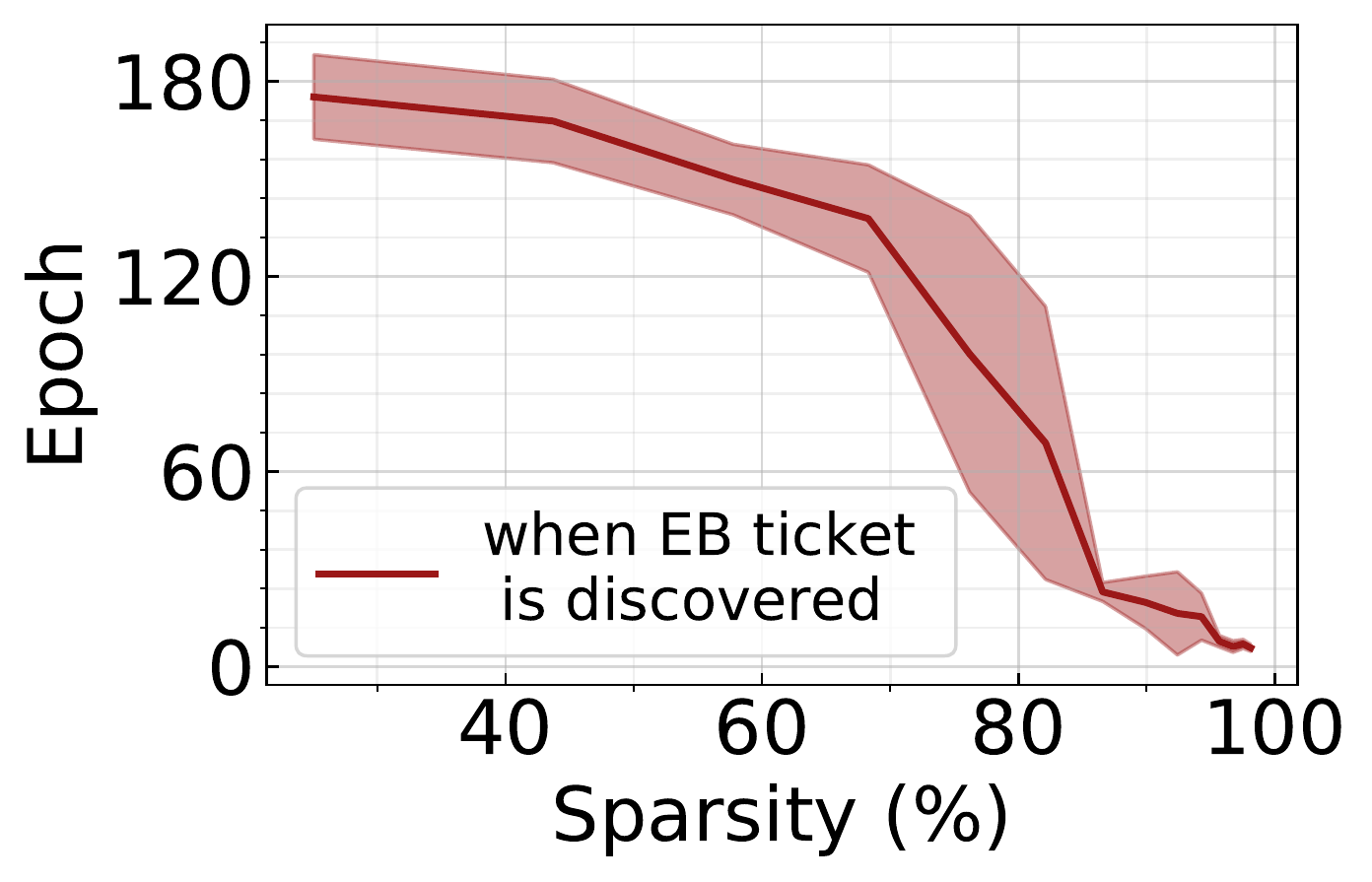} &
\includegraphics[width=0.31\linewidth]{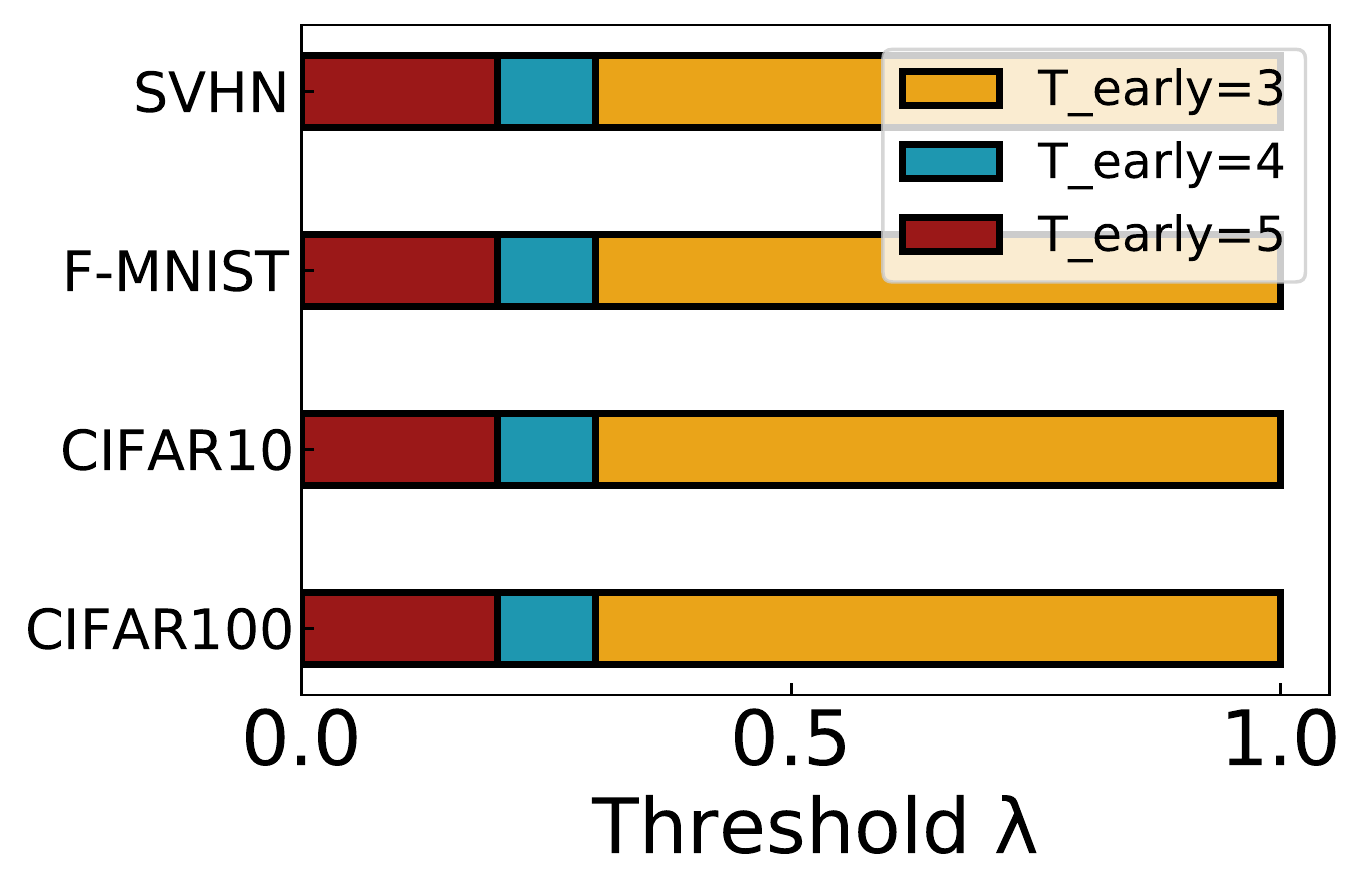}   &
\includegraphics[width=0.31\linewidth]{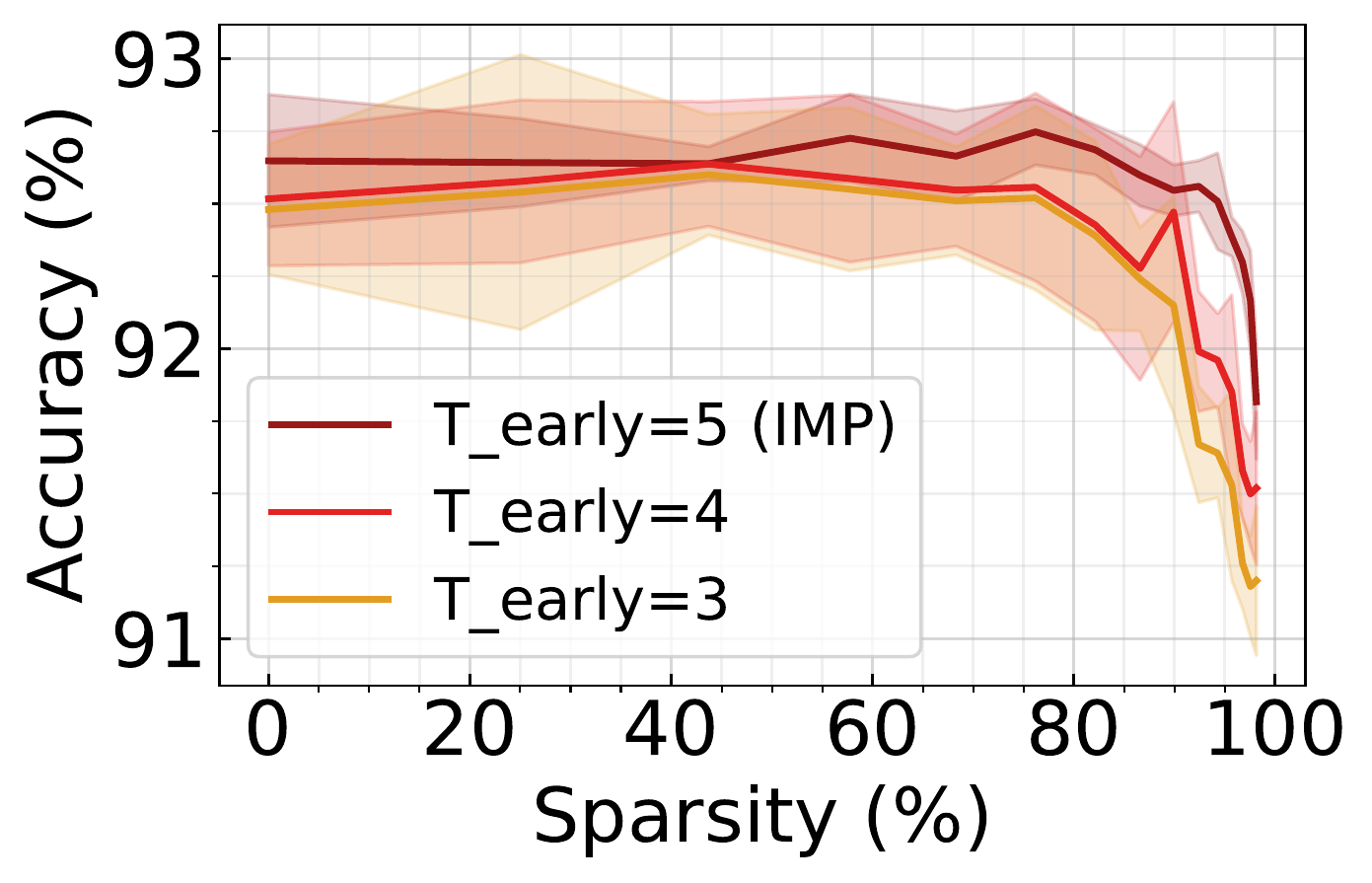}
\\
{(a)  } & { (b) }& { (c) }\\
\end{tabular}
\caption{Observations from Early-Bird (EB) ticket and Early-Time (ET) ticket.
(a) Epoch when the EB ticket is discovered. 
(b) The change of $T_{early}$ with respect to the threshold $\lambda$ for KL divergence. (c) The performance of winning tickets from different $T_{early}$. We use VGG16 on the CIFAR10 dataset and show standard deviation from 5 random runs.
}
\label{fig:exp:property_EB_ET}
\end{center}
\end{figure}

We also visualize the epoch when an EB ticket is discovered with respect to sparsity levels, in Fig. \ref{fig:exp:property_EB_ET}(a). 
The EB ticket obtains a pruning mask based on the mask difference between the current mask and the masks from the previous epochs.
Here, we observe that a highly sparse mask is discovered earlier than a lower sparsity mask according to EB ticket's mask detection algorithm (refer Fig. 3 of \cite{you2019drawing}).
Furthermore, we conduct a hyperparameter analysis of the proposed ET ticket. 
In Fig. \ref{fig:exp:property_EB_ET}(b), we show the change of $T_{early}$ with respect to the threshold $\lambda$ used for selection (Algorithm 1).
We search $\lambda$ with intervals of 0.1 from 0 to 1.
Low $\lambda$ value indicates that we select $T_{early}$ that is similar to the original timestep, and brings less efficiency gain.
Interestingly, the trend is similar across different datasets, which indicates our KL divergence works as a consistent metric.
Fig. \ref{fig:exp:property_EB_ET}(c) shows the accuracy of sparse SNNs from three different $T_{early}$ (Note, $T_{early}=5$ is the original IMP).
The results show that smaller $T_{early}$ also captures important connections in SNNs.

\subsection{Transferred Winning Tickets from ANN}

\begin{figure} [t]
    \centering
     \includegraphics[width=0.24\linewidth]{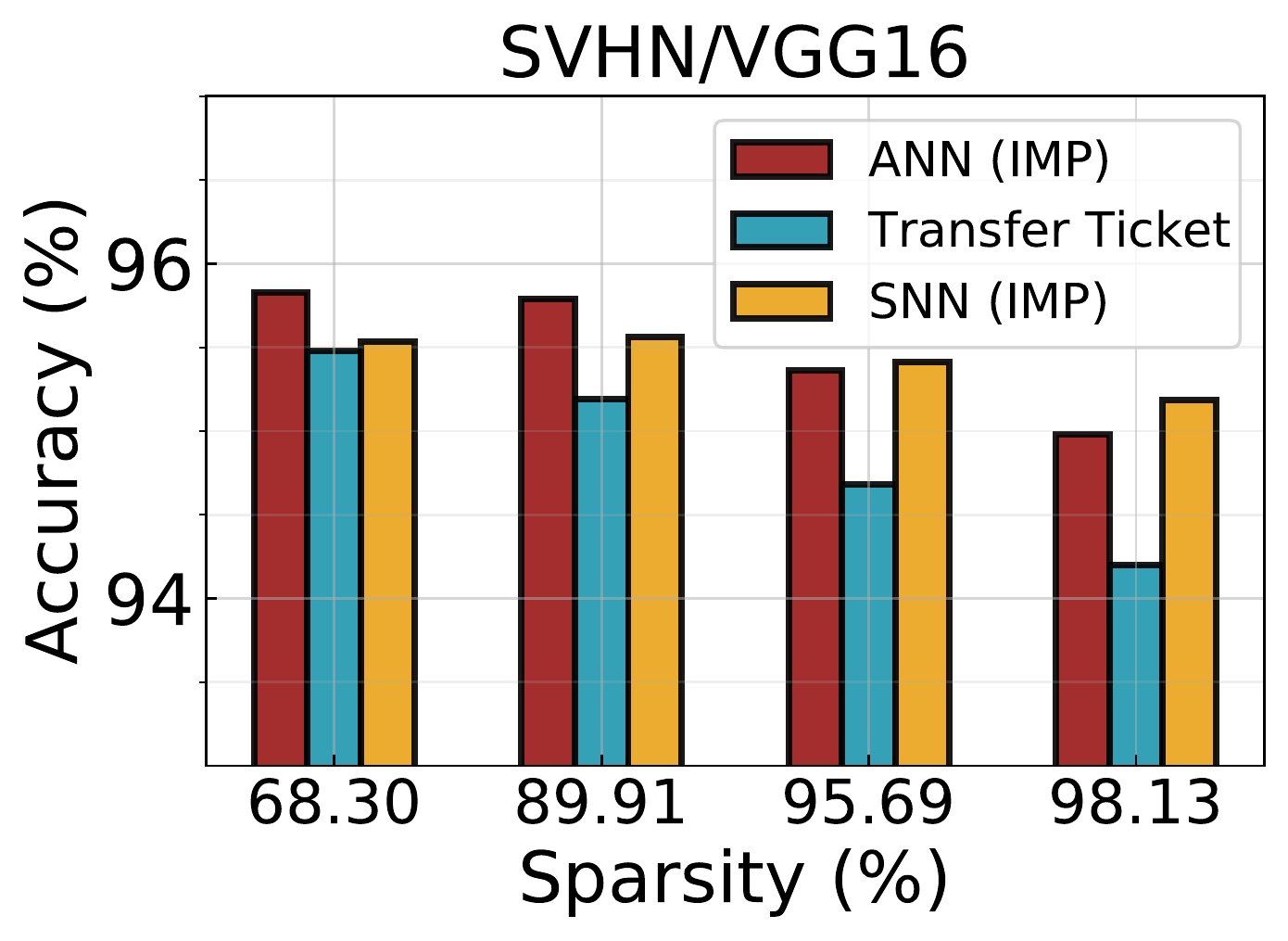}  \includegraphics[width=0.24\linewidth]{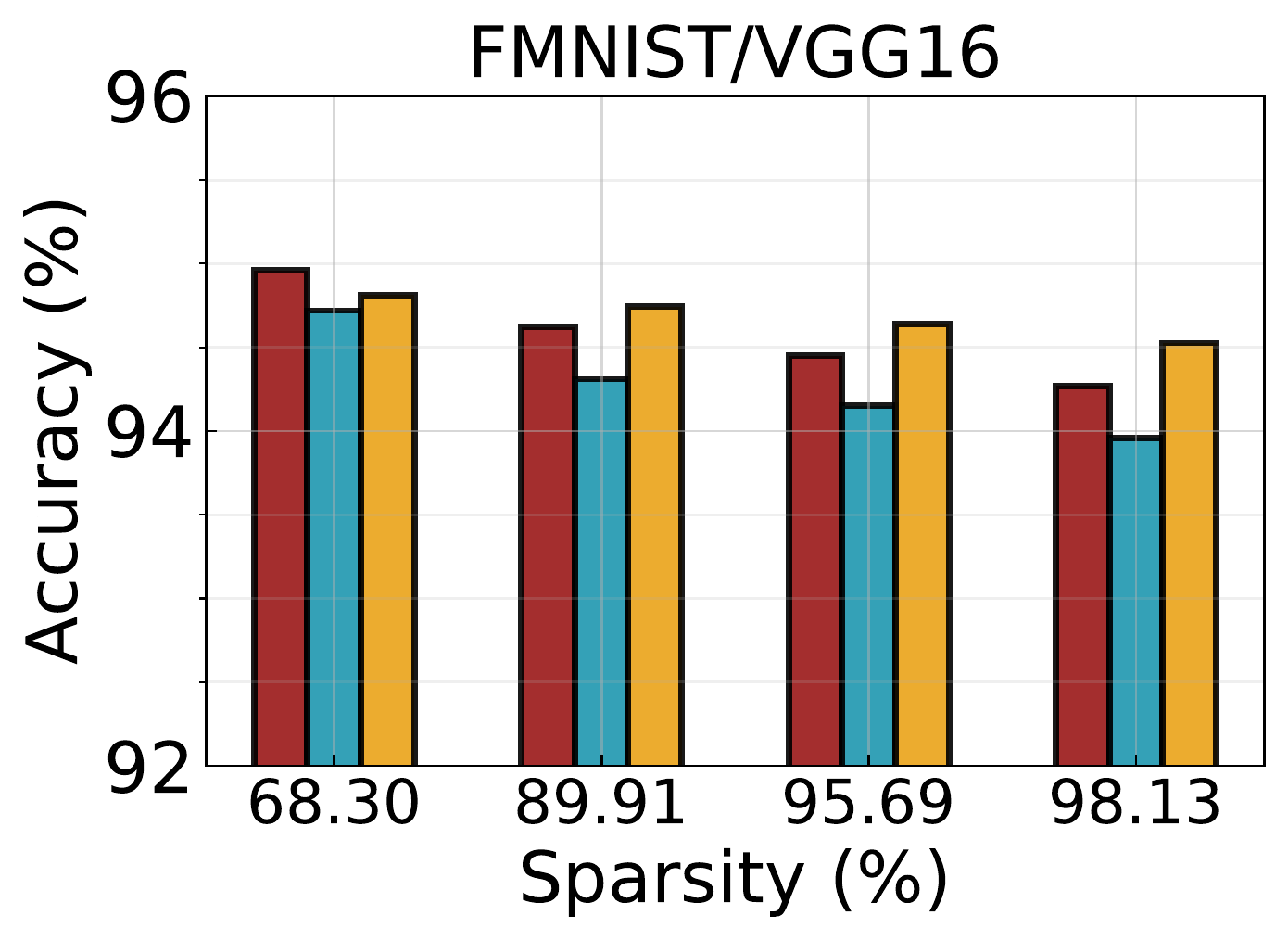}
    \includegraphics[width=0.24\linewidth]{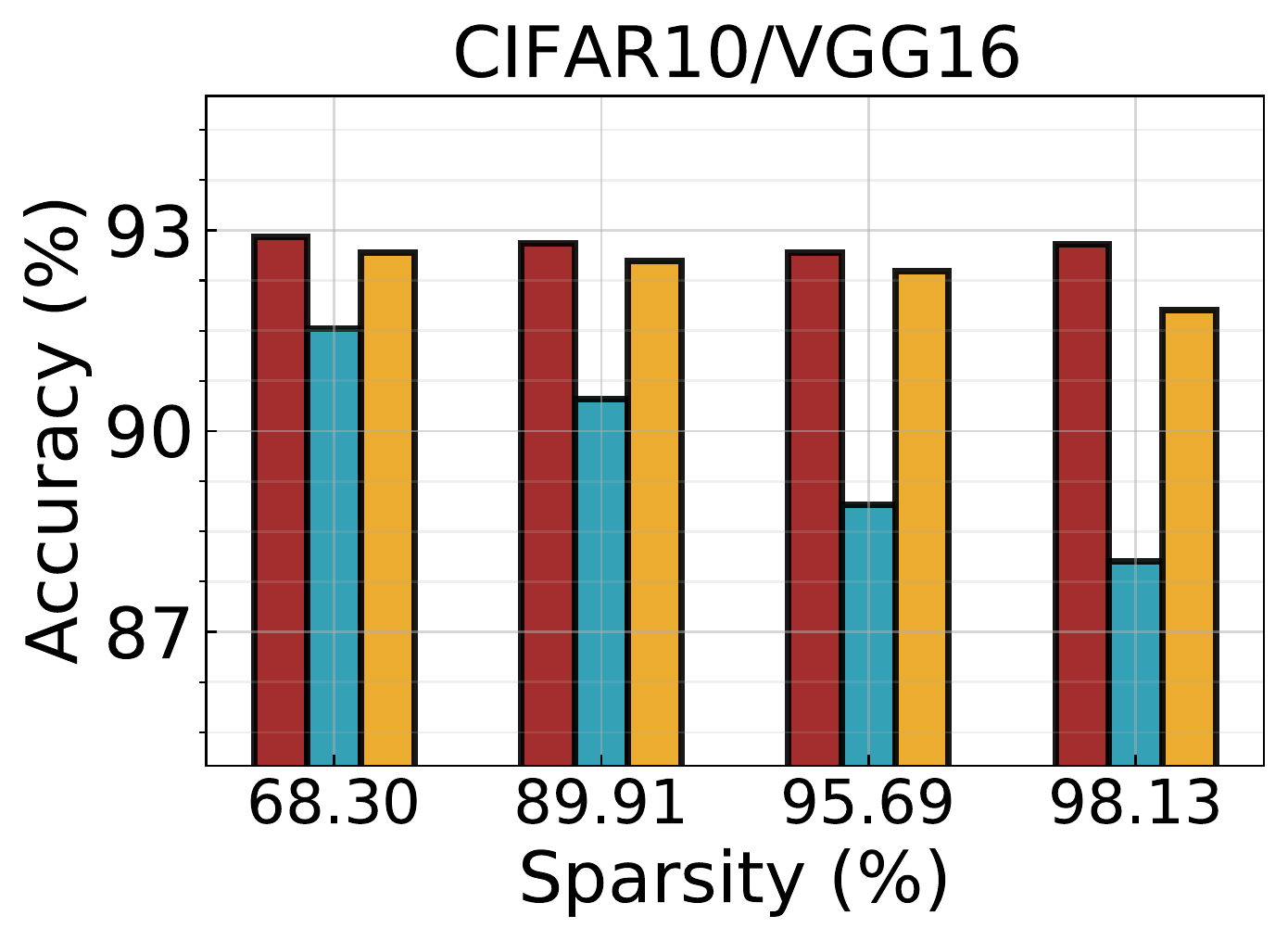}
 \includegraphics[width=0.24\linewidth]{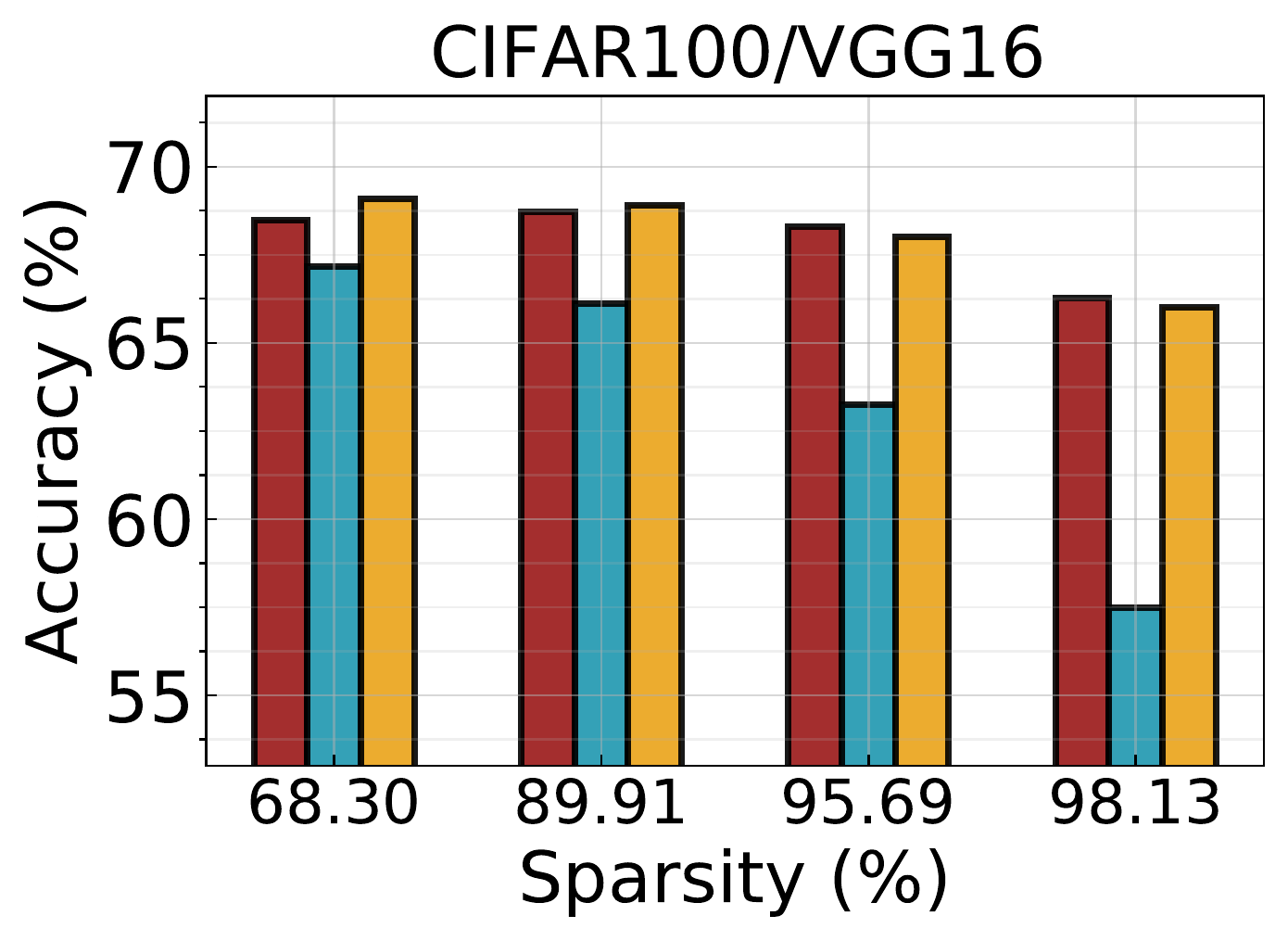} 
    \\
     \includegraphics[width=0.24\linewidth]{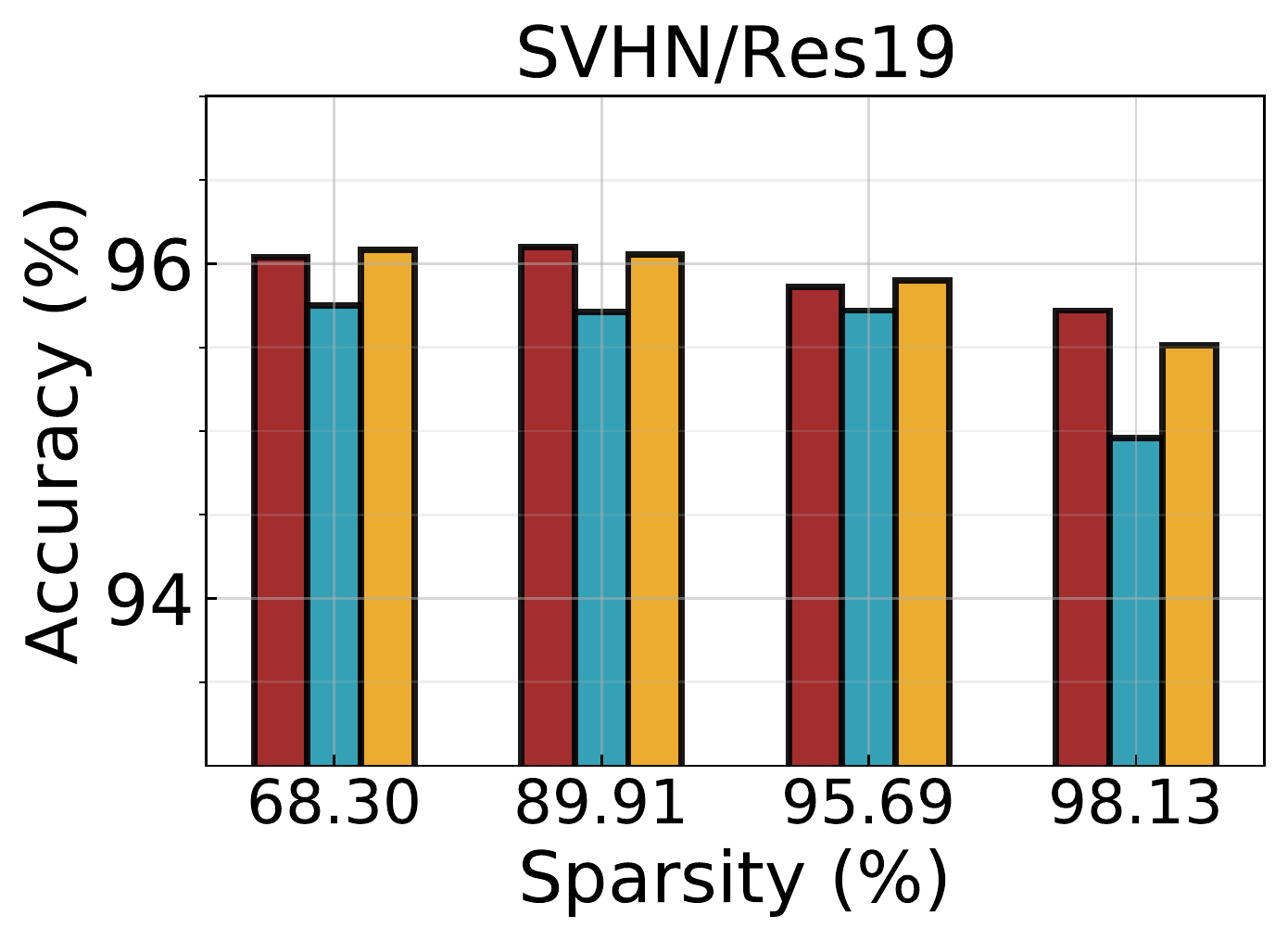}  \includegraphics[width=0.24\linewidth]{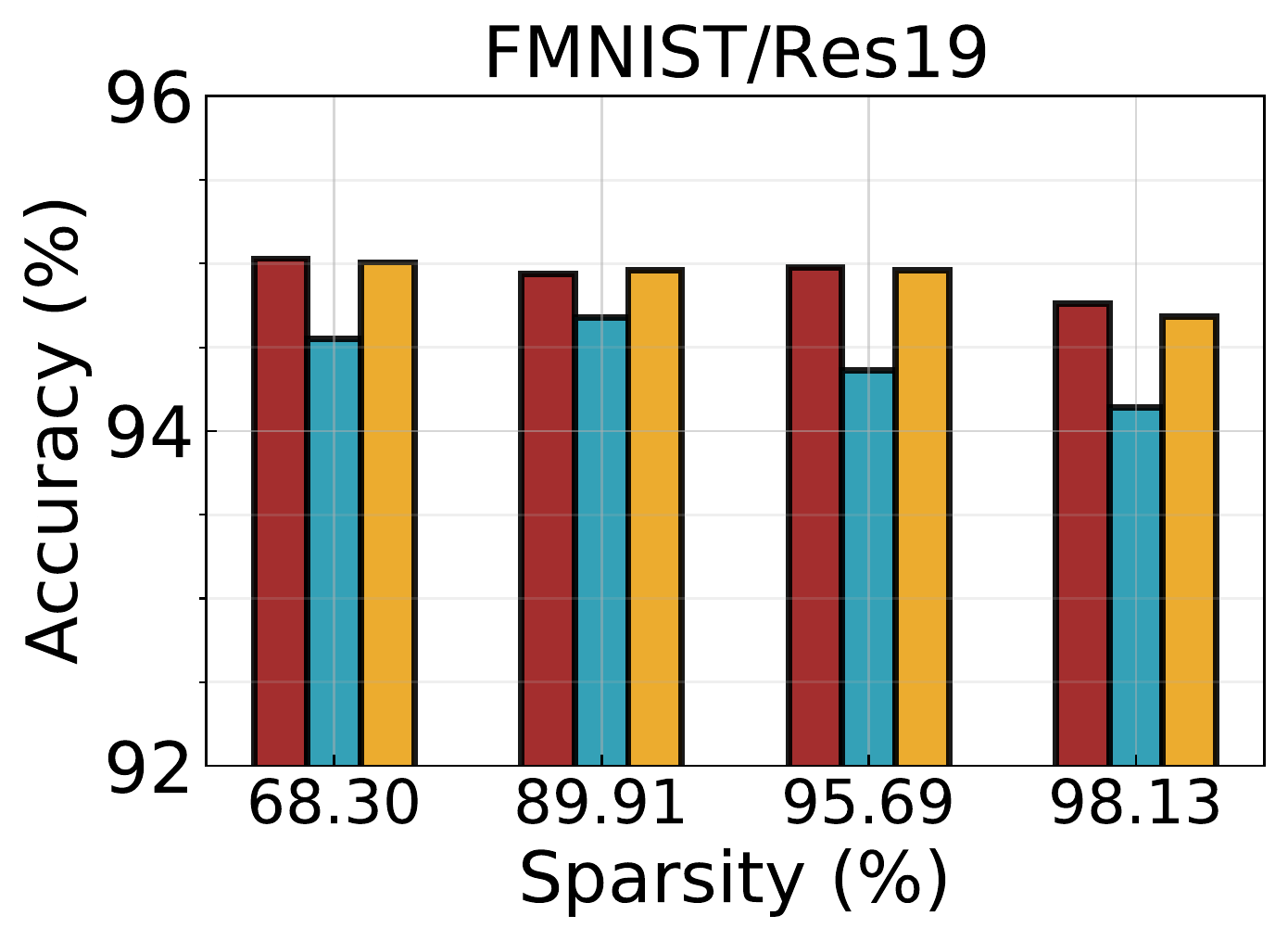}
    \includegraphics[width=0.24\linewidth]{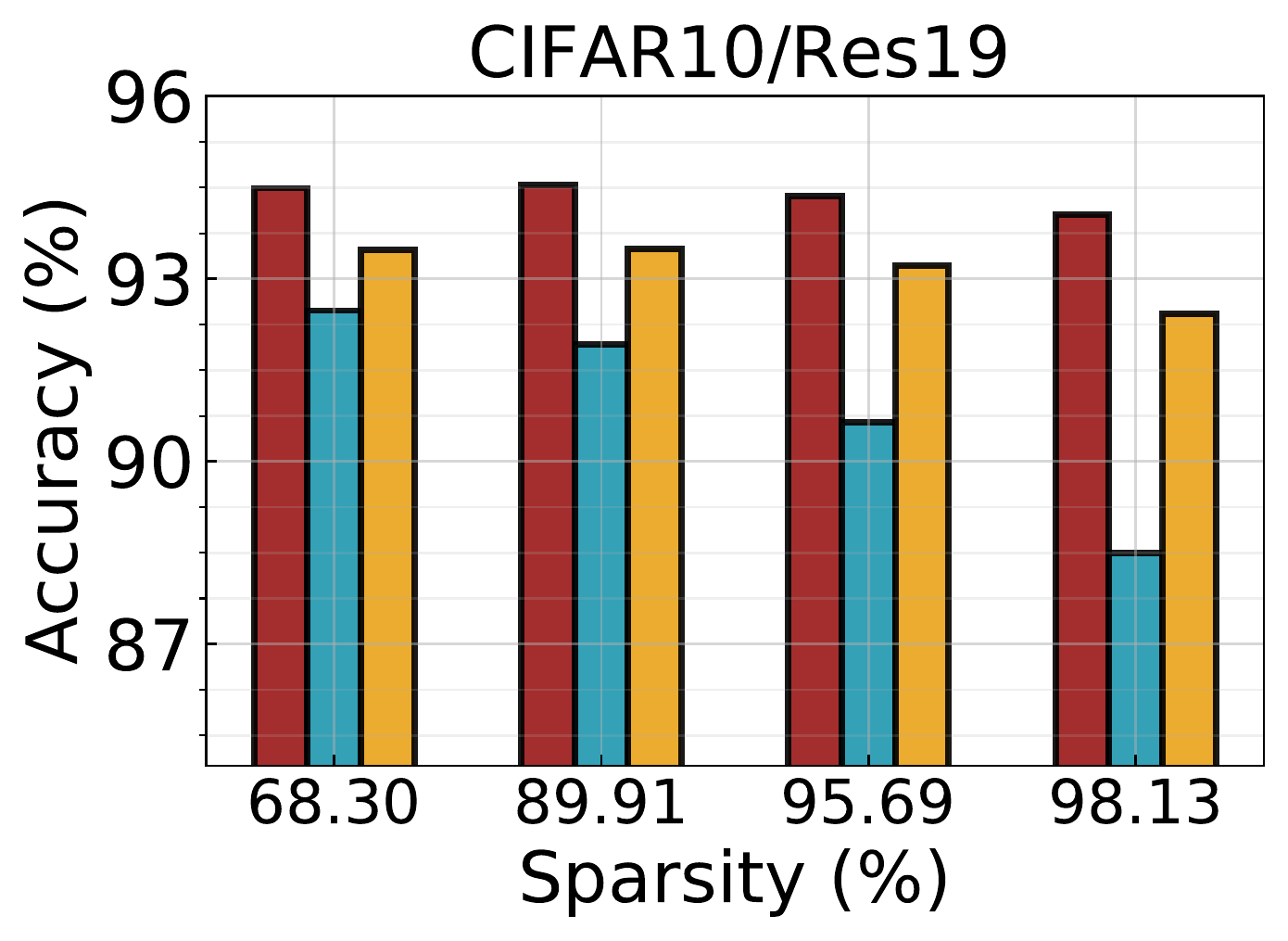}
 \includegraphics[width=0.24\linewidth]{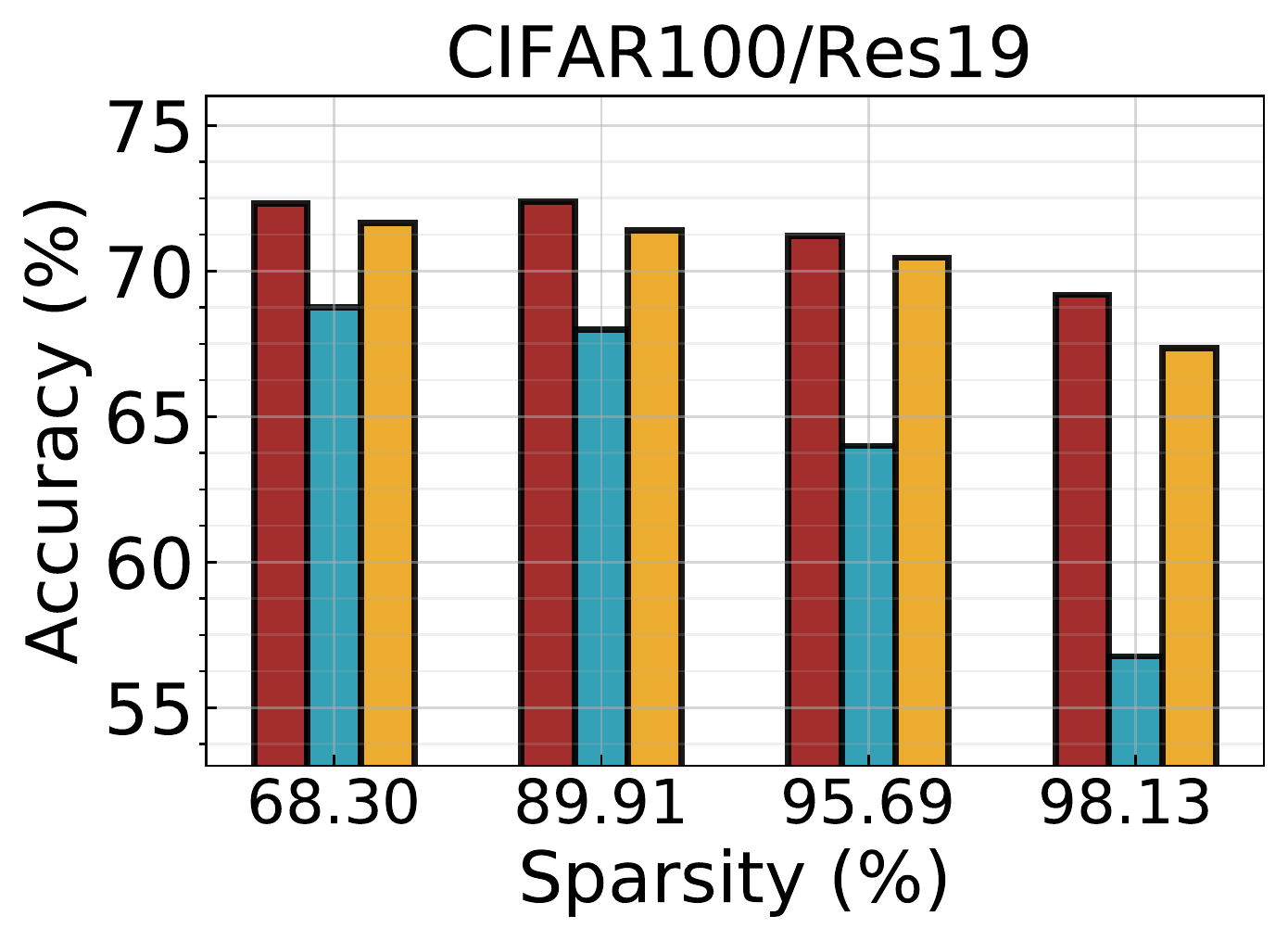} 
    \caption{Transferability study of ANN winning tickets on SNNs.}
    \label{fig:exp:transfer_ticket}
\end{figure}

The transferability of winning tickets has been actively explored in order to eliminate search costs.
A line of work \cite{mehta2019sparse,morcos2019one,desai2019evaluating,chen2020lottery} discover the existence of transferable winning tickets from the source dataset and successfully transfers it to the target dataset.
With a different perspective from the prior works which focus on cross-dataset configuration, we discover the transferable winning ticket between ANN and SNN where the activation function is different. 
In Fig. \ref{fig:exp:transfer_ticket},  we illustrate the accuracy of IMP on ANN, IMP on SNN, Transferred Ticket, across four sparsity levels (68.30\%, 89.91\%, 95.69\%, 98.13\%).
Specifically, Transferred Ticket (\ie initialized weight parameters and pruning mask) is discovered by IMP on ANN, and trained on SNN framework where we change ReLU neuron to LIF neuron.
For relatively simple datasets such as SVHN and F-MNIST, Transferred Ticket shows less than 2\% accuracy drop even at 98.13\% sparsity.      
However, for CIFAR10 and CIFAR100, Transferred Ticket fails to detect a winning ticket and shows a huge performance drop.
The results show that ANN and SNN share common knowledge, but are not exactly the same, which can be supported by the previous SNN works \cite{rathi2020enabling,ding2021optimal,kim2022privatesnn} where a pretrained ANN provides better initialization for SNN.
Although Transferred Ticket shows limited performance than IMP, searching Transferred Ticket from ANN requires $\sim 14$ hours for 98.13\% sparsity, which is $\sim 5\times$ faster than IMP on SNN.

\begin{table}[t]
\addtolength{\tabcolsep}{0.5pt}
\centering
\caption{Performance comparison of IMP \cite{frankle2018lottery} with the previous works.}
\resizebox{0.90\textwidth}{!}{%
\begin{tabular}{ccccccccc}
\toprule
\:\:Pruning Method\:\: &\: \:Architecture\: \: &\:\:\:\: \:Dataset\: \: &\:\: Baseline Acc. ($\%$)  \:\:&\:\: $\Delta$Acc ($\%$) \:\: & \:\: Sparsity ($\%$)\:\:\\
\midrule
\multirow{3}{*}{Deng \etal \cite{deng2021comprehensive}} 
 & \multirow{3}{*}{7Conv, 2FC} & \multirow{3}{*}{CIFAR10}  & \multirow{3}{*}{89.53} &  -0.38 & 50.00  \\ & & &  &  -2.16   &75.00  \\ &  &  & &  -3.85 &90.00    \\
 \midrule
\multirow{3}{*}{Bellec \etal \cite{bellec2018long}} 
 & \multirow{3}{*}{6Conv, 2FC}& \multirow{3}{*}{CIFAR10}   &  \multirow{3}{*}{92.84}  & -1.98  & 94.76 \\ &  &  & &   -2.56 & 98.05  \\ &  &  &  &   -3.53  & 98.96 \\
 \midrule
\multirow{3}{*}{Chen \etal \cite{chen2021pruning}} 
 & \multirow{3}{*}{6Conv, 2FC}& \multirow{3}{*}{CIFAR10}   &  \multirow{3}{*}{92.84} &   -0.30  & 71.59 \\ &  &  &   &  -0.81 & 94.92  \\ &  &   &  &   -1.47  & 97.65 \\
  \midrule
\multirow{3}{*}{Chen \etal  \cite{chen2021pruning}$^*$} 
 & \multirow{3}{*}{ResNet19}& \multirow{3}{*}{CIFAR10}   &  \multirow{3}{*}{93.22} &   -0.54  & 76.90 \\ &  &  &  &   -1.31 & 94.25  \\ &  &  &   &   -2.10  & 97.56 \\
    \midrule
\multirow{3}{*}{\textbf{IMP  on SNN (ours)} } 
 & \multirow{3}{*}{\textbf{ResNet19}}& \multirow{3}{*}{CIFAR10}   &  \multirow{3}{*}{\textbf{93.22}} &   \textbf{+0.28}  &  \textbf{76.20} \\ &  &  &   &  \textbf{+0.24} & \textbf{94.29}  \\ &  &   &   &  \textbf{-0.04}  &\textbf{97.54} \\
  \midrule
   \midrule
 \multirow{3}{*}{Chen \etal \cite{chen2021pruning}$^*$} 
 & \multirow{3}{*}{ResNet19}& \multirow{3}{*}{CIFAR100}   &  \multirow{3}{*}{71.34} &   -1.98 & 77.03 \\ &  &  &  &   -3.87 & 94.92  \\ &  &  &   &   -4.03  & 97.65 \\
    \midrule
\multirow{3}{*}{\textbf{IMP  on SNN (ours)} } 
 & \multirow{3}{*}{\textbf{ResNet19}}& \multirow{3}{*}{CIFAR100}   &  \multirow{3}{*}{\textbf{71.34}} &   \textbf{+0.11}  &  \textbf{76.20} \\ &  &  &   &  \textbf{-0.34} & \textbf{94.29}  \\ &  &   &   &  \textbf{-2.29}  &\textbf{97.54} \\
\bottomrule
 \multicolumn{6}{l}{$^*$ We reimplement ResNet19 experiments.  }
\end{tabular}%
}
\label{table:exp:comparison_with_previous}
\end{table}

\subsection{Finding Winning Tickets from Initialization}

\begin{figure}[t]
\noindent\begin{minipage}{\textwidth}
\begin{minipage}{0.48\textwidth}
\centering
\includegraphics[width=0.75\linewidth]{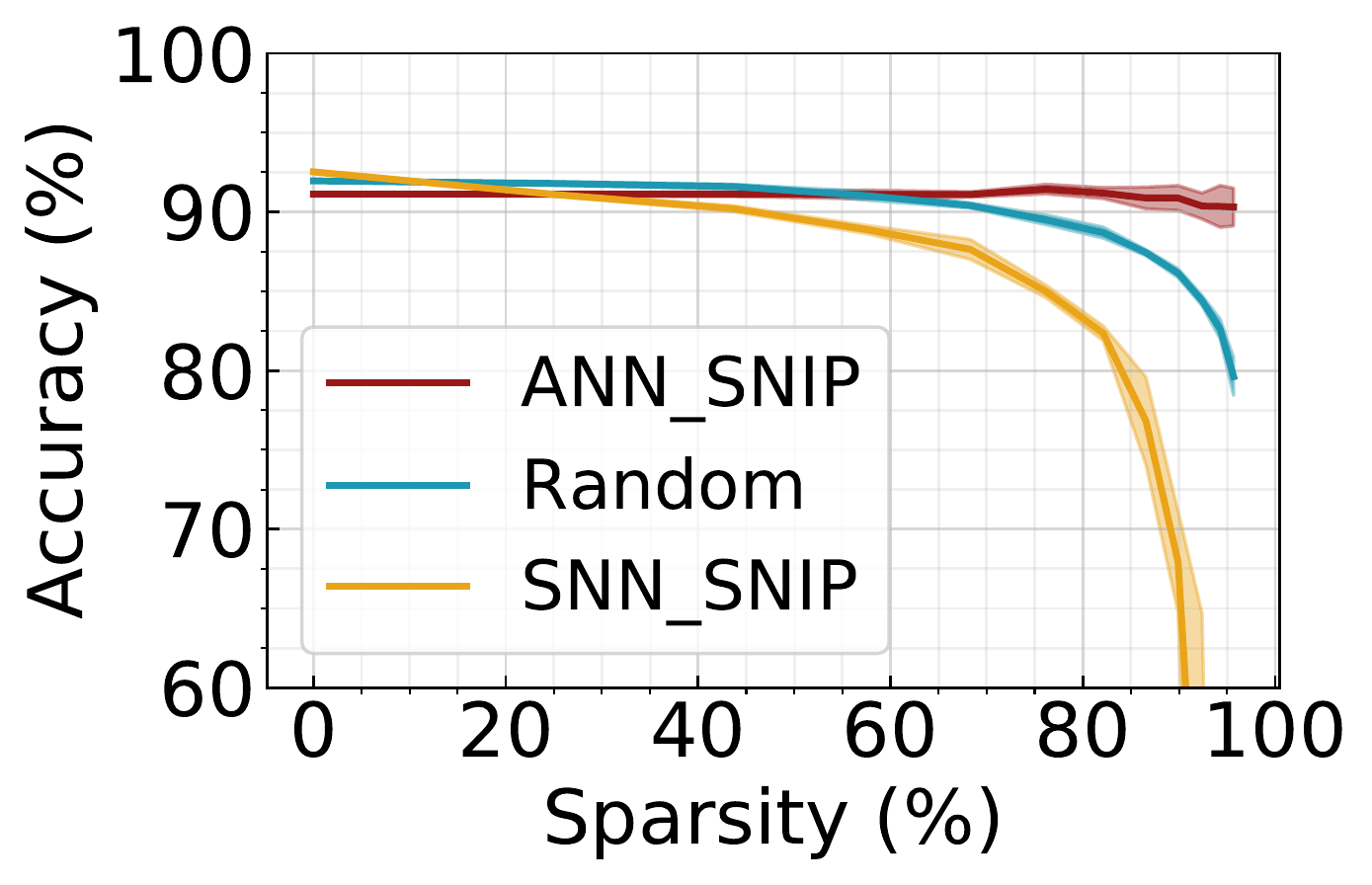}
\caption{Performance comparison across pruned ANN and pruned SNN with SNIP.}
    \label{fig:exp:snip}
\end{minipage}
\hfill
\begin{minipage}{0.48\textwidth}
\centering
\includegraphics[width=0.75\linewidth]{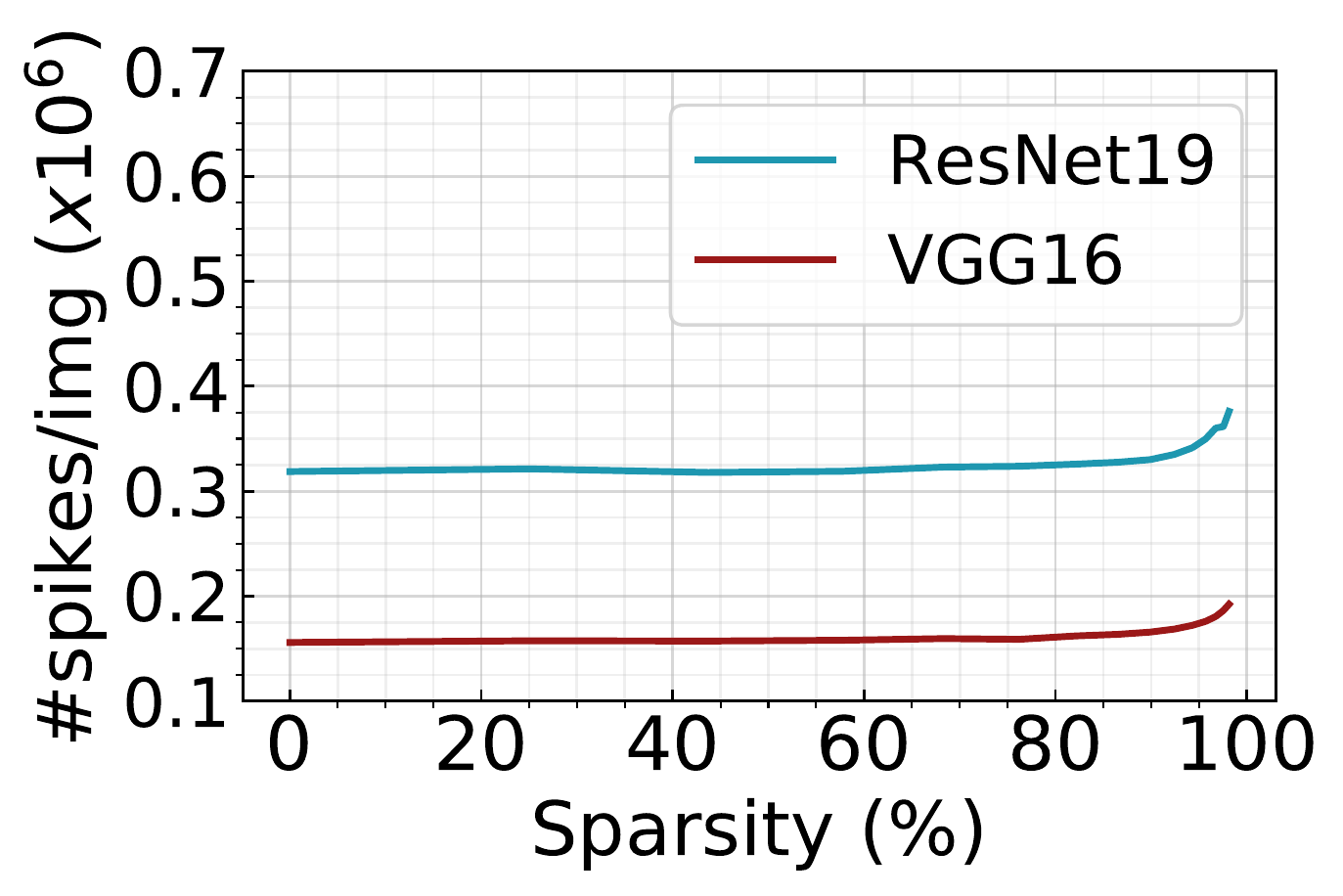}
\caption{Number of spikes with respect to sparsity on CIFAR10.}
 \label{fig:exp:sparsity}
\end{minipage}
\end{minipage}
\end{figure}

A line of work \cite{lee2018snip,wang2020picking} effectively reduces search cost for winning tickets by conducting a search process at initialization.
This technique should be explored with SNNs where multiple feedforward steps corresponding to multiple timesteps bring expensive search costs.
To show this, we conduct experiments on a representative \textit{pruning at initialization} method, SNIP \cite{lee2018snip}.
SNIP computes the importance of each weight connection from the magnitude of backward gradients at initialization.
In Fig. \ref{fig:exp:snip}, we illustrate the accuracy of ANN and SNN with SNIP on VGG16/CIFAR10 configuration. 

Surprisingly, SNN shows huge performance degradation at high sparsity regime ($>80\%$), even worse than random pruning.
The results imply that the previous \textit{pruning at initialization} based on the backward gradient (tailored for ANNs) is not compatible with SNNs where the backward gradient is approximated because of the non-differentiability of LIF neuron \cite{neftci2019surrogate,wu2018spatio}.

\subsection{Performance Comparison with Previous Works}

In Table \ref{table:exp:comparison_with_previous}, we compare the LTH method, especially IMP, with state-of-the-art SNN pruning works \cite{deng2021comprehensive,bellec2018long,chen2021pruning} in terms of accuracy and achieved sparsity.
We show the baseline accuracy of the unpruned model and report the accuracy drop at three different sparsity levels. Note that the previous works use a shallow model with $6\sim7$ conv and 2 FC layers.
The ResNet19 architecture achieves higher baseline accuracy compared to the previous shallow architectures.
To compare the results on a ResNet19 model, we prune ResNet19 with method proposed by Chen \etal \cite{chen2021pruning} using the official code\footnote[1]
{https://github.com/Yanqi-Chen/Gradient-Rewiring} provided by authors.
We observe that the previous SNN pruning works fail to achieve matching performance at a high sparsity on deeper SNN architectures.
On the other hand, IMP shows less performance drop (in some cases even performance improvement) compared to the previous pruning techniques.
Especially, at $97.54\%$ sparsity, \cite{chen2021pruning} shows $2.1\%$ accuracy drop whereas IMP degrades the accuracy by $0.04\%$.
Further, we compare the accuracy on the CIFAR100 dataset to explore the effectiveness of the pruning method with respect to complex datasets.
Chen \etal's method fails to discover the winning ticket at sparsity 94.29\%, and IMP shows better performance across all sparsity levels.
The results imply that the LTH-based method might bring a huge advantage as SNNs are scaled up in the future.

\subsection{Observation on the Number of Spikes}

In general, the energy consumption of SNNs is proportional to the number of spikes \cite{akopyan2015truenorth,davies2018loihi,yin2022sata} and weight sparsity.
In Fig. \ref{fig:exp:sparsity}, we measure the number of spikes per image across various sparsity levels of VGG16 and ResNet19 architectures. We use IMP for pruning SNNs. 
We observe that the sparse SNN maintains a similar number of spikes across all sparsity levels.
The results indicate that sparse SNNs obtained with LTH do not bring an additional MAC energy-efficiency gain from spike sparsity.
Nonetheless, weight sparsity brings less memory occupation and can alleviate the memory movement overheads \cite{chen2016eyeriss,parashar2017scnn}. 
As discussed in \cite{weight_sparsity_paper1}, with over $98\%$ of the sparse weights, SNNs can reduce the memory movement energy by up to $15 \times$.

\section{Conclusion}

In this work, we aim to explore lottery ticket hypothesis based pruning for implementing sparse deep SNNs at lower search costs.
Such an objective is important as SNNs are a promising candidate for deployment on resource-constrained edge devices.
To this end, we apply various techniques including Iterative Magnitude Pruning (IMP), Early-Bird ticket (EB), and the proposed Early-Time ticket (ET).  Our key observations are summarized as follows:
(1) IMP can achieve higher sparsity levels of deep SNN models compared to previous works on SNN pruning. 
(2) EB ticket reduces search time significantly, but it cannot achieve a winning ticket over $90\%$ sparsity.
(3) Adding ET ticket accelerates search speed for both IMP and EB, by up to $\times 1.66$. We can find a winning ticket in one hour with EB+ET, which can enable practical pruning on edge devices.

\noindent{\textbf{Acknowledgement.}}  We would like to thank Anna Hambitzer for her helpful comments.
This  work  was  supported  in  part  by  C-BRIC,  a  JUMP center sponsored by DARPA and SRC, Google Research Scholar Award, the National Science Foundation (Grant\#1947826), TII (Abu Dhabi) and the DARPA AI Exploration (AIE) program.

%
%
\bibliographystyle{splncs04}
\bibliography{egbib,egbib_snnpruning,egbib_annpruning}

\begin{thebibliography}{10}
\providecommand{\url}[1]{\texttt{#1}}
\providecommand{\urlprefix}{URL }
\providecommand{\doi}[1]{https://doi.org/#1}

\bibitem{akopyan2015truenorth}
Akopyan, F., Sawada, J., Cassidy, A., Alvarez-Icaza, R., Arthur, J., Merolla,
  P., Imam, N., Nakamura, Y., Datta, P., Nam, G.J., et~al.: Truenorth: Design
  and tool flow of a 65 mw 1 million neuron programmable neurosynaptic chip.
  IEEE transactions on computer-aided design of integrated circuits and systems
   \textbf{34}(10),  1537--1557 (2015)

\bibitem{bai2022dual}
Bai, Y., Wang, H., TAO, Z., Li, K., Fu, Y.: Dual lottery ticket hypothesis. In:
  International Conference on Learning Representations (2022),
  \url{https://openreview.net/forum?id=fOsN52jn25l}

\bibitem{bellec2018long}
Bellec, G., Salaj, D., Subramoney, A., Legenstein, R., Maass, W.: Long
  short-term memory and learning-to-learn in networks of spiking neurons.
  Advances in neural information processing systems  \textbf{31} (2018)

\bibitem{brix2020successfully}
Brix, C., Bahar, P., Ney, H.: Successfully applying the stabilized lottery
  ticket hypothesis to the transformer architecture. arXiv preprint
  arXiv:2005.03454  (2020)

\bibitem{burkholz2021existence}
Burkholz, R., Laha, N., Mukherjee, R., Gotovos, A.: On the existence of
  universal lottery tickets. arXiv preprint arXiv:2111.11146  (2021)

\bibitem{weight_sparsity_paper1}
Chen, G.K., Kumar, R., Sumbul, H.E., Knag, P.C., Krishnamurthy, R.K.: A
  4096-neuron 1m-synapse 3.8-pj/sop spiking neural network with on-chip stdp
  learning and sparse weights in 10-nm finfet cmos. IEEE Journal of Solid-State
  Circuits  \textbf{54}(4),  992--1002 (2018)

\bibitem{chen2020lottery}
Chen, T., Frankle, J., Chang, S., Liu, S., Zhang, Y., Wang, Z., Carbin, M.: The
  lottery ticket hypothesis for pre-trained bert networks. Advances in neural
  information processing systems  \textbf{33},  15834--15846 (2020)

\bibitem{chen2022sparsity}
Chen, T., Zhang, Z., pengjun wang, Balachandra, S., Ma, H., Wang, Z., Wang, Z.:
  Sparsity winning twice: Better robust generalization from more efficient
  training. In: International Conference on Learning Representations (2022),
  \url{https://openreview.net/forum?id=SYuJXrXq8tw}

\bibitem{chen2021pruning}
Chen, Y., Yu, Z., Fang, W., Huang, T., Tian, Y.: Pruning of deep spiking neural
  networks through gradient rewiring. arXiv preprint arXiv:2105.04916  (2021)

\bibitem{chen2016eyeriss}
Chen, Y.H., Emer, J., Sze, V.: Eyeriss: A spatial architecture for
  energy-efficient dataflow for convolutional neural networks. ACM SIGARCH
  Computer Architecture News  \textbf{44}(3),  367--379 (2016)

\bibitem{christensen20222022}
Christensen, D.V., Dittmann, R., Linares-Barranco, B., Sebastian, A., Le~Gallo,
  M., Redaelli, A., Slesazeck, S., Mikolajick, T., Spiga, S., Menzel, S.,
  et~al.: 2022 roadmap on neuromorphic computing and engineering. Neuromorphic
  Computing and Engineering  (2022)

\bibitem{comsa2020temporal}
Comsa, I.M., Fischbacher, T., Potempa, K., Gesmundo, A., Versari, L.,
  Alakuijala, J.: Temporal coding in spiking neural networks with alpha
  synaptic function. In: ICASSP 2020-2020 IEEE International Conference on
  Acoustics, Speech and Signal Processing (ICASSP). pp. 8529--8533. IEEE (2020)

\bibitem{davies2018loihi}
Davies, M., Srinivasa, N., Lin, T.H., Chinya, G., Cao, Y., Choday, S.H., Dimou,
  G., Joshi, P., Imam, N., Jain, S., et~al.: Loihi: A neuromorphic manycore
  processor with on-chip learning. IEEE Micro  \textbf{38}(1),  82--99 (2018)

\bibitem{deng2021comprehensive}
Deng, L., Wu, Y., Hu, Y., Liang, L., Li, G., Hu, X., Ding, Y., Li, P., Xie, Y.:
  Comprehensive snn compression using admm optimization and activity
  regularization. IEEE transactions on neural networks and learning systems
  (2021)

\bibitem{deng2022temporal}
Deng, S., Li, Y., Zhang, S., Gu, S.: Temporal efficient training of spiking
  neural network via gradient re-weighting. In: International Conference on
  Learning Representations (2022),
  \url{https://openreview.net/forum?id=_XNtisL32jv}

\bibitem{desai2019evaluating}
Desai, S., Zhan, H., Aly, A.: Evaluating lottery tickets under distributional
  shifts. arXiv preprint arXiv:1910.12708  (2019)

\bibitem{ding2021optimal}
Ding, J., Yu, Z., Tian, Y., Huang, T.: Optimal ann-snn conversion for fast and
  accurate inference in deep spiking neural networks. arXiv preprint
  arXiv:2105.11654  (2021)

\bibitem{ding2022audio}
Ding, S., Chen, T., Wang, Z.: Audio lottery: Speech recognition made
  ultra-lightweight, noise-robust, and transferable. In: International
  Conference on Learning Representations (2022),
  \url{https://openreview.net/forum?id=9Nk6AJkVYB}

\bibitem{SpikingJelly}
Fang, W., Chen, Y., Ding, J., Chen, D., Yu, Z., Zhou, H., Tian, Y., other
  contributors: Spikingjelly.
  \url{https://github.com/fangwei123456/spikingjelly} (2020)

\bibitem{fang2021deep}
Fang, W., Yu, Z., Chen, Y., Huang, T., Masquelier, T., Tian, Y.: Deep residual
  learning in spiking neural networks. Advances in Neural Information
  Processing Systems  \textbf{34} (2021)

\bibitem{frankle2018lottery}
Frankle, J., Carbin, M.: The lottery ticket hypothesis: Finding sparse,
  trainable neural networks. arXiv preprint arXiv:1803.03635  (2018)

\bibitem{frankle2019stabilizing}
Frankle, J., Dziugaite, G.K., Roy, D.M., Carbin, M.: Stabilizing the lottery
  ticket hypothesis. arXiv preprint arXiv:1903.01611  (2019)

\bibitem{furber2014spinnaker}
Furber, S.B., Galluppi, F., Temple, S., Plana, L.A.: The spinnaker project.
  Proceedings of the IEEE  \textbf{102}(5),  652--665 (2014)

\bibitem{girish2021lottery}
Girish, S., Maiya, S.R., Gupta, K., Chen, H., Davis, L.S., Shrivastava, A.: The
  lottery ticket hypothesis for object recognition. In: Proceedings of the
  IEEE/CVF Conference on Computer Vision and Pattern Recognition. pp. 762--771
  (2021)

\bibitem{guo2020unsupervised}
Guo, W., Fouda, M.E., Yantir, H.E., Eltawil, A.M., Salama, K.N.: Unsupervised
  adaptive weight pruning for energy-efficient neuromorphic systems. Frontiers
  in Neuroscience p.~1189 (2020)

\bibitem{han2016dsd}
Han, S., Pool, J., Narang, S., Mao, H., Gong, E., Tang, S., Elsen, E., Vajda,
  P., Paluri, M., Tran, J., et~al.: Dsd: Dense-sparse-dense training for deep
  neural networks. arXiv preprint arXiv:1607.04381  (2016)

\bibitem{han2015learning}
Han, S., Pool, J., Tran, J., Dally, W.: Learning both weights and connections
  for efficient neural network. Advances in neural information processing
  systems  \textbf{28} (2015)

\bibitem{he2016deep}
He, K., Zhang, X., Ren, S., Sun, J.: Deep residual learning for image
  recognition. In: CVPR. pp. 770--778 (2016)

\bibitem{ioffe2015batch}
Ioffe, S., Szegedy, C.: Batch normalization: Accelerating deep network training
  by reducing internal covariate shift. arXiv preprint arXiv:1502.03167  (2015)

\bibitem{izhikevich2003simple}
Izhikevich, E.M.: Simple model of spiking neurons. IEEE Transactions on neural
  networks  \textbf{14}(6),  1569--1572 (2003)

\bibitem{kalibhat2020winning}
Kalibhat, N.M., Balaji, Y., Feizi, S.: Winning lottery tickets in deep
  generative models. arXiv preprint arXiv:2010.02350  (2020)

\bibitem{kim2022neural}
Kim, Y., Li, Y., Park, H., Venkatesha, Y., Panda, P.: Neural architecture
  search for spiking neural networks. arXiv preprint arXiv:2201.10355  (2022)

\bibitem{kim2020revisiting}
Kim, Y., Panda, P.: Revisiting batch normalization for training low-latency
  deep spiking neural networks from scratch. Frontiers in neuroscience p.~1638
  (2020)

\bibitem{kim2021visual}
Kim, Y., Panda, P.: Visual explanations from spiking neural networks using
  interspike intervals. Sci Rep 11, 19037 (2021).
  https://doi.org/10.1038/s41598-021-98448-0  (2021)

\bibitem{kim2022privatesnn}
Kim, Y., Venkatesha, Y., Panda, P.: Privatesnn: Privacy-preserving spiking
  neural networks. In: Proceedings of the AAAI Conference on Artificial
  Intelligence. vol.~36, pp. 1192--1200 (2022)

\bibitem{krizhevsky2009learning}
Krizhevsky, A., Hinton, G., et~al.: Learning multiple layers of features from
  tiny images  (2009)

\bibitem{kundu2021hire}
Kundu, S., Pedram, M., Beerel, P.A.: Hire-snn: Harnessing the inherent
  robustness of energy-efficient deep spiking neural networks by training with
  crafted input noise. In: Proceedings of the IEEE/CVF International Conference
  on Computer Vision. pp. 5209--5218 (2021)

\bibitem{ledinauskas2020training}
Ledinauskas, E., Ruseckas, J., Jur{\v{s}}{\.e}nas, A., Bura{\v{c}}as, G.:
  Training deep spiking neural networks. arXiv preprint arXiv:2006.04436
  (2020)

\bibitem{lee2020enabling}
Lee, C., Sarwar, S.S., Panda, P., Srinivasan, G., Roy, K.: Enabling spike-based
  backpropagation for training deep neural network architectures. Frontiers in
  Neuroscience  \textbf{14} (2020)

\bibitem{lee2016training}
Lee, J.H., Delbruck, T., Pfeiffer, M.: Training deep spiking neural networks
  using backpropagation. Frontiers in neuroscience  \textbf{10}, ~508 (2016)

\bibitem{lee2018snip}
Lee, N., Ajanthan, T., Torr, P.H.: Snip: Single-shot network pruning based on
  connection sensitivity. arXiv preprint arXiv:1810.02340  (2018)

\bibitem{li2016pruning}
Li, H., Kadav, A., Durdanovic, I., Samet, H., Graf, H.P.: Pruning filters for
  efficient convnets. arXiv preprint arXiv:1608.08710  (2016)

\bibitem{li2021free}
Li, Y., Deng, S., Dong, X., Gong, R., Gu, S.: A free lunch from ann: Towards
  efficient, accurate spiking neural networks calibration. arXiv preprint
  arXiv:2106.06984  (2021)

\bibitem{li2022converting}
Li, Y., Deng, S., Dong, X., Gu, S.: Converting artificial neural networks to
  spiking neural networks via parameter calibration. arXiv preprint
  arXiv:2205.10121  (2022)

\bibitem{li2021differentiable}
Li, Y., Guo, Y., Zhang, S., Deng, S., Hai, Y., Gu, S.: Differentiable spike:
  Rethinking gradient-descent for training spiking neural networks. Advances in
  Neural Information Processing Systems  \textbf{34} (2021)

\bibitem{liu2021deep}
Liu, S., Chen, T., Atashgahi, Z., Chen, X., Sokar, G., Mocanu, E., Pechenizkiy,
  M., Wang, Z., Mocanu, D.C.: Deep ensembling with no overhead for either
  training or testing: The all-round blessings of dynamic sparsity. arXiv
  preprint arXiv:2106.14568  (2021)

\bibitem{liu2018rethinking}
Liu, Z., Sun, M., Zhou, T., Huang, G., Darrell, T.: Rethinking the value of
  network pruning. arXiv preprint arXiv:1810.05270  (2018)

\bibitem{loshchilov2016sgdr}
Loshchilov, I., Hutter, F.: Sgdr: Stochastic gradient descent with warm
  restarts. arXiv preprint arXiv:1608.03983  (2016)

\bibitem{martinelli2020spiking}
Martinelli, F., Dellaferrera, G., Mainar, P., Cernak, M.: Spiking neural
  networks trained with backpropagation for low power neuromorphic
  implementation of voice activity detection. In: ICASSP 2020-2020 IEEE
  International Conference on Acoustics, Speech and Signal Processing (ICASSP).
  pp. 8544--8548. IEEE (2020)

\bibitem{mehta2019sparse}
Mehta, R.: Sparse transfer learning via winning lottery tickets. arXiv preprint
  arXiv:1905.07785  (2019)

\bibitem{morcos2019one}
Morcos, A., Yu, H., Paganini, M., Tian, Y.: One ticket to win them all:
  generalizing lottery ticket initializations across datasets and optimizers.
  Advances in neural information processing systems  \textbf{32} (2019)

\bibitem{mostafa2017supervised}
Mostafa, H.: Supervised learning based on temporal coding in spiking neural
  networks. IEEE transactions on neural networks and learning systems
  \textbf{29}(7),  3227--3235 (2017)

\bibitem{movva2020dissecting}
Movva, R., Zhao, J.Y.: Dissecting lottery ticket transformers: Structural and
  behavioral study of sparse neural machine translation. arXiv preprint
  arXiv:2009.13270  (2020)

\bibitem{neftci2019surrogate}
Neftci, E.O., Mostafa, H., Zenke, F.: Surrogate gradient learning in spiking
  neural networks. IEEE Signal Processing Magazine  \textbf{36},  61--63 (2019)

\bibitem{neftci2016stochastic}
Neftci, E.O., Pedroni, B.U., Joshi, S., Al-Shedivat, M., Cauwenberghs, G.:
  Stochastic synapses enable efficient brain-inspired learning machines.
  Frontiers in neuroscience  \textbf{10}, ~241 (2016)

\bibitem{netzer2011reading}
Netzer, Y., Wang, T., Coates, A., Bissacco, A., Wu, B., Ng, A.Y.: Reading
  digits in natural images with unsupervised feature learning  (2011)

\bibitem{orchard2021efficient}
Orchard, G., Frady, E.P., Rubin, D.B.D., Sanborn, S., Shrestha, S.B., Sommer,
  F.T., Davies, M.: Efficient neuromorphic signal processing with loihi 2. In:
  2021 IEEE Workshop on Signal Processing Systems (SiPS). pp. 254--259. IEEE
  (2021)

\bibitem{parashar2017scnn}
Parashar, A., Rhu, M., Mukkara, A., Puglielli, A., Venkatesan, R., Khailany,
  B., Emer, J., Keckler, S.W., Dally, W.J.: Scnn: An accelerator for
  compressed-sparse convolutional neural networks. ACM SIGARCH computer
  architecture news  \textbf{45}(2),  27--40 (2017)

\bibitem{paszke2017automatic}
Paszke, A., Gross, S., Chintala, S., Chanan, G., Yang, E., DeVito, Z., Lin, Z.,
  Desmaison, A., Antiga, L., Lerer, A.: Automatic differentiation in pytorch.
  In: NIPS-W (2017)

\bibitem{rathi2018stdp}
Rathi, N., Panda, P., Roy, K.: Stdp-based pruning of connections and weight
  quantization in spiking neural networks for energy-efficient recognition.
  IEEE Transactions on Computer-Aided Design of Integrated Circuits and Systems
   \textbf{38}(4),  668--677 (2018)

\bibitem{rathi2021diet}
Rathi, N., Roy, K.: Diet-snn: A low-latency spiking neural network with direct
  input encoding and leakage and threshold optimization. IEEE Transactions on
  Neural Networks and Learning Systems  (2021)

\bibitem{rathi2020enabling}
Rathi, N., Srinivasan, G., Panda, P., Roy, K.: Enabling deep spiking neural
  networks with hybrid conversion and spike timing dependent backpropagation.
  arXiv preprint arXiv:2005.01807  (2020)

\bibitem{roy2019towards}
Roy, K., Jaiswal, A., Panda, P.: Towards spike-based machine intelligence with
  neuromorphic computing. Nature  \textbf{575}(7784),  607--617 (2019)

\bibitem{schuman2022opportunities}
Schuman, C.D., Kulkarni, S.R., Parsa, M., Mitchell, J.P., Kay, B., et~al.:
  Opportunities for neuromorphic computing algorithms and applications. Nature
  Computational Science  \textbf{2}(1),  10--19 (2022)

\bibitem{shi2019soft}
Shi, Y., Nguyen, L., Oh, S., Liu, X., Kuzum, D.: A soft-pruning method applied
  during training of spiking neural networks for in-memory computing
  applications. Frontiers in neuroscience  \textbf{13}, ~405 (2019)

\bibitem{shrestha2018slayer}
Shrestha, S.B., Orchard, G.: Slayer: Spike layer error reassignment in time.
  arXiv preprint arXiv:1810.08646  (2018)

\bibitem{simonyan2014very}
Simonyan, K., Zisserman, A.: Very deep convolutional networks for large-scale
  image recognition. ICLR  (2015)

\bibitem{venkatesha2021federated}
Venkatesha, Y., Kim, Y., Tassiulas, L., Panda, P.: Federated learning with
  spiking neural networks. arXiv preprint arXiv:2106.06579  (2021)

\bibitem{vischer2021lottery}
Vischer, M.A., Lange, R.T., Sprekeler, H.: On lottery tickets and minimal task
  representations in deep reinforcement learning. arXiv preprint
  arXiv:2105.01648  (2021)

\bibitem{wang2020picking}
Wang, C., Zhang, G., Grosse, R.: Picking winning tickets before training by
  preserving gradient flow. arXiv preprint arXiv:2002.07376  (2020)

\bibitem{wen2016learning}
Wen, W., Wu, C., Wang, Y., Chen, Y., Li, H.: Learning structured sparsity in
  deep neural networks. Advances in neural information processing systems
  \textbf{29} (2016)

\bibitem{wu2021training}
Wu, H., Zhang, Y., Weng, W., Zhang, Y., Xiong, Z., Zha, Z.J., Sun, X., Wu, F.:
  Training spiking neural networks with accumulated spiking flow. ijo
  \textbf{1}(1) (2021)

\bibitem{wu2020progressive}
Wu, J., Xu, C., Zhou, D., Li, H., Tan, K.C.: Progressive tandem learning for
  pattern recognition with deep spiking neural networks. arXiv preprint
  arXiv:2007.01204  (2020)

\bibitem{wu2018spatio}
Wu, Y., Deng, L., Li, G., Zhu, J., Shi, L.: Spatio-temporal backpropagation for
  training high-performance spiking neural networks. Frontiers in neuroscience
  \textbf{12}, ~331 (2018)

\bibitem{wu2019direct}
Wu, Y., Deng, L., Li, G., Zhu, J., Xie, Y., Shi, L.: Direct training for
  spiking neural networks: Faster, larger, better. In: Proceedings of the AAAI
  Conference on Artificial Intelligence. vol.~33, pp. 1311--1318 (2019)

\bibitem{wu2022brain}
Wu, Y., Zhao, R., Zhu, J., Chen, F., Xu, M., Li, G., Song, S., Deng, L., Wang,
  G., Zheng, H., et~al.: Brain-inspired global-local learning incorporated with
  neuromorphic computing. Nature Communications  \textbf{13}(1),  1--14 (2022)

\bibitem{xiao2017fashion}
Xiao, H., Rasul, K., Vollgraf, R.: Fashion-mnist: a novel image dataset for
  benchmarking machine learning algorithms. arXiv preprint arXiv:1708.07747
  (2017)

\bibitem{yao2021temporal}
Yao, M., Gao, H., Zhao, G., Wang, D., Lin, Y., Yang, Z., Li, G.: Temporal-wise
  attention spiking neural networks for event streams classification. In:
  Proceedings of the IEEE/CVF International Conference on Computer Vision. pp.
  10221--10230 (2021)

\bibitem{yin2022sata}
Yin, R., Moitra, A., Bhattacharjee, A., Kim, Y., Panda, P.: Sata:
  Sparsity-aware training accelerator for spiking neural networks. arXiv
  preprint arXiv:2204.05422  (2022)

\bibitem{you2019drawing}
You, H., Li, C., Xu, P., Fu, Y., Wang, Y., Chen, X., Baraniuk, R.G., Wang, Z.,
  Lin, Y.: Drawing early-bird tickets: Towards more efficient training of deep
  networks. arXiv preprint arXiv:1909.11957  (2019)

\bibitem{yu2019playing}
Yu, H., Edunov, S., Tian, Y., Morcos, A.S.: Playing the lottery with rewards
  and multiple languages: lottery tickets in rl and nlp. arXiv preprint
  arXiv:1906.02768  (2019)

\bibitem{zhang2021efficient}
Zhang, Z., Chen, X., Chen, T., Wang, Z.: Efficient lottery ticket finding: Less
  data is more. In: International Conference on Machine Learning. pp.
  12380--12390. PMLR (2021)

\bibitem{zheng2020going}
Zheng, H., Wu, Y., Deng, L., Hu, Y., Li, G.: Going deeper with directly-trained
  larger spiking neural networks. arXiv preprint arXiv:2011.05280  (2020)

\bibitem{zhou2019deconstructing}
Zhou, H., Lan, J., Liu, R., Yosinski, J.: Deconstructing lottery tickets:
  Zeros, signs, and the supermask. Advances in neural information processing
  systems  \textbf{32} (2019)

\end{thebibliography}

\end{document}